\documentclass{article}
\usepackage{ijcai21}

\RequirePackage[T1]{fontenc} 
\RequirePackage[tt=false, type1=true]{libertine} 
\RequirePackage[varqu]{zi4} 

\RequirePackage[libertine]{newtxmath}
\usepackage[letterspace=-25]{microtype}

\usepackage{natbib}
\usepackage{soul}
\usepackage{url}
\usepackage[hidelinks]{hyperref}
\usepackage[utf8]{inputenc}
\usepackage[small]{caption}
\usepackage{graphicx}
\usepackage{amsmath}
\usepackage{booktabs}
\usepackage{savesym}
\savesymbol{Bbbk}
\savesymbol{openbox}

\usepackage{amssymb}
\usepackage{amsthm}
\usepackage{MnSymbol}
\usepackage{color}
\usepackage{url}
\usepackage{multirow}
\usepackage{xcolor}
\usepackage{dsfont}
\usepackage{algorithm}
\usepackage{algpseudocode}

\restoresymbol{NEW}{Bbbk}
\restoresymbol{NEW}{openbox}

\usepackage[switch]{lineno}

\usepackage{tablefootnote}

\usepackage{subcaption}

\usepackage{todonotes}
\usepackage{xargs}
\usepackage{xcolor}

\usepackage{multicol}
\usepackage{multirow}
\usepackage{makecell}
\usepackage{xargs}
\usepackage{xcolor}
\usepackage{hyperref}

\def\rational{coherent}
\def\rationality{coherence}
\def\irrational{incoherent}
\def\Rational{Coherent}
\def\Rationality{Coherence}

\graphicspath{{images}}

\def\name{ArguCast}
\def\sname{ACF}

\newcommand{\dg}[1]{\textcolor{orange}{#1}}

\newtheorem{definition}{Definition}
\newtheorem{property}{Property}
\newtheorem{proposition}{Proposition}

\newcommand{\AR}[1]{\textcolor{blue}{#1}}
\newcommand{\FT}[1]{\textcolor{red}{#1}}
\newcommand{\note}[1]{\textcolor{magenta}{$*$ #1}}
\newcommand{\delete}[1]{\textcolor{olive}{$\times$ #1}}

\newcounter{RQ}

\newtheorem{example}{Example}

\title{
Argumentatively Coherent Judgmental Forecasting\\ (with Supplementary Material)}

\author{Deniz Gorur$^1$ \and Antonio Rago$^{1,2}$ \and Francesca Toni$^1$ \\
  $^1$ Department of Computing, Imperial College London, UK \\
  $^2$ Department of Informatics, King's College London, UK \\
  \texttt{\{d.gorur22,a.rago,ft\}@imperial.ac.uk}
  }

\begin{document}
\maketitle







\begin{abstract}
\emph{Judgmental forecasting} employs 
human opinions 
to make predictions about future events, rather than exclusively 
historical data as in quantitative forecasting. 
When these opinions form an argumentative structure around forecasts, it is useful to study the properties of the forecasts from an argumentative perspective. In this paper, we advocate and formally define a property of argumentative \emph{\rationality},
which, in essence, requires that a forecaster's reasoning is \rational\ with their forecast.
We then conduct three evaluations 
with our notion of \rationality.
First, we
assess the impact of enforcing \rationality\ on human forecasters as well as on Large Language Model (LLM)-based forecasters, given that they have recently shown to be competitive with human forecasters. In both cases, we show that 
filtering out \irrational\ predictions improves forecasting accuracy consistently, supporting the practical value of \rationality\ in both human and LLM-based forecasting.
Then, 
via crowd-sourced user experiments, we show that, despite its apparent intuitiveness and usefulness,
users do not generally align with this \rationality\ property
.
This points to the need to integrate, within argumentation-based judgmental forecasting, mechanisms to filter out \irrational\ opinions before obtaining group forecasting predictions
.

\end{abstract}

\section{Introduction}

Judgmental forecasting involves the use of human opinion 
to predict future events, as opposed to relying solely on historical data 
as in quantitative forecasting \cite{lawrence2006judgmental,zellner2021human}. In the area of judgmental forecasting, \citet{tetlock2014forecasting_tournaments} established techniques 
for supporting forecasting
, and \citet{tetlock2016superforecasting} improved these techniques by selecting ``Superforecasters'' (forecasters who naturally excel at making predictions).
However, judgmental forecasting can be influenced by irrationality and cognitive biases \cite{kahneman2011thinking}, which can lead to inaccurate predictions. The importance of \rationality\ in this context cannot be overstated \cite{karvetski2022forecasting_rationales}. For instance, if a forecaster allows their own personal biases to influence their prediction, it could lead to skewed, and thus less accurate, results. 

To overcome irrationality and biases, many solutions have been proposed over the years, e.g. \cite{wallstenB1995ReviewLinguistic,arkes2001overconfidence,welton2016developingExpert,zellner2021human}
,
%
including some solutions
\cite{karvetskiOMT2013ProbCoherence,fan2019improving,gibsonMRO2021ImprovingProbabilisic,thomson2019combining,ho2024measuring} showing that imposing coherence improves forecasting accuracy.
Further, 
\rationality\ constraints has been shown to help in hierarchical forecasting \cite{rangapuramWBMGJ2021CoherentProbabilisticForecasts,panagiotelisGAH2023ProbabilisticReconciliation,rangapuramKNMJ2023CoherentProbabilistic,kin2024ProbabilisticCoherentAgg}.
%
Differently from these approaches, 
in this paper we 
propose a novel solution based on a notion of \rationality\ 
deployed 
alongside 
debates (on forecasting questions) 
structured as in \textit{computational argumentation} (see \cite{AImagazine17,handbook} for overviews). This field of research has emerged in recent years in AI as a methodology to conjugate representational needs 
of user-related cognitive models with computational models for automated reasoning \cite{lippi2016ArgumentationMiningState}. These computational models are formalised as \emph{argumentation frameworks}, which excel in reasoning with 
incomplete information and for resolving conflicts.
Within this landscape, existing research \cite{irwin2022forecasting,gorur2023argucast,toni2023forecasting} attempts to address the aforementioned issues with judgmental forecasting
with the use of 
argumentation frameworks
. 
Some of these works~\cite{toni2023forecasting} do not incorporate any \rationality\ constraints.
Other work~\cite{irwin2022forecasting}, while showing promising results with the introduction of \rationality\ constraints, focus on restricted settings, notably they can only handle binary questions and disallow 
the same arguments to be used for different questions, which does not maximise the potential of argumentation.
Further work~\cite{gorur2023argucast} 
{addresses these limitations 
while imposing \rationality\ constraints,
but shows limitations as regards structuring and evaluating forecasters' judgments (for example, forecasters may express no opinion but their judgment may be different from neutral).   
}

In this paper,
we make several contributions.

\begin{itemize}
    \item     
    We define a novel, (arguably) natural notion of \rational\ forecaster behaviour
    , given predictions and votes on arguments, for  
    a variant of the forecasting method of \cite{gorur2023argucast} {addressing its limitations as regards structuring and evaluating forecasters’ judgments.
    Differently from existing notions, our notion of \rationality\ is adaptable to different forecasting contexts through controllable parameters.
    }
    
    %
    \item Then, to assess the usefulness of our notion of \rationality\ in practice, we 
    carry out three evaluations.
    \begin{enumerate} 
        \item 
        We conduct an in-person user study with a 
        variant of the 
        argumentation-based forecasting system ArguCast\footnote{\href{http://argucast.herokuapp.com}{argucast.herokuapp.com}} 
        {(this variant implements our variant of the forecasting method of \cite{gorur2023argucast}).}
        With this study, we aim to see whether applying \rationality\ gets  predictions closer to 
        state-of-the-art forecasting systems.
        \item We perform experiments on Large Language Model (LLM)-based forecasters, given that they have recently shown to be competitive with human forecasters \cite{halawi2024approaching}. We use Argumentative LLMs (ArgLLMs)~\cite{freedman2024argllms} to first generate arguments for and against the forecasting arguments, and then we apply \rationality\ to see whether forecasting accuracy improves.
        \item 
        We run a crowd-sourced user study where we assign users variants of debates, formulated 
        with our forecasting method, and assess 
        their alignment with our notion of \rationality.
    \end{enumerate}
\end{itemize}
Results from the in-person user study and from the experiments with the LLMs show that applying \rationality\ provides consistent improvements on forecasting accuracy, thus showing the practical need for \rationality\ constraints.
Results from the crowd-sourced study demonstrate that users only partially align with our notion. Thus, given the benefits of \rationality\ towards accuracy, this final 
study
points to the need to enforce \rationality\ constraints for filtering out \irrational\ opinions in \name, and potentially also in other argumentation-based forecasting tools (e.g. \cite{irwin2022forecasting}).  


This pre-print is an extended version of \cite{ecaiversion}, including supplementary material.

\section{Related Work}
\label{sec:related_work}
\paragraph{Judgmental Forecasting Baselines}
We measure our method, filtering out the opinion of \irrational\ forecasters, against a 
baselines representing the state-of-the-art in (judgmental) forecasting. These are Metaculus\footnote{\href{https://www.metaculus.com/}{www.metaculus.com}} and prediction markets (in particular we use Betfair\footnote{\href{https://www.betfair.com/}{www.betfair.com}}).

\paragraph{\Rationality\ in Judgmental Forecasting}
In
Judgmental forecasting, there is some work that tries to incorporate \rationality, for example by
investigating the best ways of eliciting probabilities from humans \cite{wallstenB1995ReviewLinguistic}, how incentives and training change forecasters' abilities \cite{welton2016developingExpert}, the effect of scoring rules on forecasters \cite{zellner2021human}, and how principles help overcome overconfidence \cite{arkes2001overconfidence}.
Some 
existing solutions \cite{karvetskiOMT2013ProbCoherence,fan2019improving,gibsonMRO2021ImprovingProbabilisic} 
show that forecasting accuracy improves by weighting judges' assessments based on their degree of coherence.
Further, \cite{thomson2019combining} develop a measure of \rationality\ to identify the degree of independence between sets of forecasts, thus improving the aggregated forecasts. Additionally, \cite{ho2024measuring} introduce a notion of probabilistic coherence that identifies superior forecasters. Finally, there has been some work on \rationality\ constraints in hierarchical forecasting \cite{rangapuramWBMGJ2021CoherentProbabilisticForecasts,panagiotelisGAH2023ProbabilisticReconciliation,rangapuramKNMJ2023CoherentProbabilistic,kin2024ProbabilisticCoherentAgg}.
In this paper we do not compare our notion against existing approaches, since they do not operate in our setting.


\paragraph{Forecasting and Argumentation}
Existing research \cite{irwin2022forecasting,gorur2023argucast,toni2023forecasting} explores the integration of computational argumentation for eliciting, representing and evaluating the argumentative reasoning of users.
Specifically, \citet{irwin2022forecasting} were the first to apply computational argumentation to support judgmental forecasting.
They introduced their notion of \rationality\ based on how forecasters update their predictions and their reasoning behind those updates.
However, as mentioned in the introduction, their underlying framework exhibits some limitations in that it can only handle binary questions and does not allow the same arguments to be used for different questions. 
It is also important to note that our notion of \rationality\ is based solely on forecasters' argumentative reasoning and their predictions rather than 
prediction updates.
Further demonstrating the benefits of argumentation in this context is \cite{toni2023forecasting}, which incorporates users' opinions and probabilistic estimates to aid forecasting, but without considering \rationality\ constraints.
In addition, the role of argumentation 
for forecasting is indirectly advocated by resources such as
the FORECAST2023 dataset \cite{GorskaL2024forecast2023}, where the rationales from the Hybrid-Forecasting Competition 
are annotated with support/attack relations.

We use a variant of ArguCast \cite{gorur2023argucast}, which addressed the shortcomings of \cite{irwin2022forecasting}
.
Our variant 
exhibits additional properties (e.g. neutral and bias-based forecasting) for the underlying formalism for structuring and evaluating forecasters' judgments 
(see Appendix~\ref{app:proofs}).
Also, our notion of \rationality\ is parameterised and adaptable to different forecasting contexts, whereas the notion of \rationality\ in \cite{gorur2023argucast} is fixed.

\paragraph{LLMs and 
Forecasting}
Recent research demonstrates the potential of LLMs 
in forecasting, e.g. in time-series forecasting, LLMs have been shown to improve accuracy \cite{jin2023time,jin2024time,liu2024taming}. However, direct use of LLMs in judgmental forecasting has yielded mixed results: while some studies \cite{schoenegger2023large,abolghasemi2023humans} show that using these models directly does not improve forecasting accuracy, others \cite{halawi2024approaching} have concluded that the current LLMs are nearly as good as human forecasters if they are supported with Retrieval Augmented Generation (RAG) \cite{Zhao2024RAGSurvey}. They further point that the aggregation of 
forecasts from humans and LLMs with RAG achieve the best forecasting accuracy.
Combining human insights with LLM-supported advice has also proven effective in accuracy 
\cite{schoenegger2024aiaugmented}.
Our approach in Section~\ref{sec:empirical_exp} incorporates LLMs on forecaster data to improve forecasting accuracy.
It does so by adapting, towards forecasting, ArgLLMs~\cite{freedman2024argllms}, an existing system for mining argumentation frameworks from LLMs 
to verify claims.

\section{Background}
\label{sec:background}
In this section, we 
give 
the relevant background on computational argumentation, including for forecasting.

A \textbf{Quantitative Bipolar Argumentation Framework (QBAF)} \cite{baroni2019gradual} is a tuple $\langle\mathcal{X}, \mathcal{A}, \mathcal{S}, \tau\rangle$ where 
$\mathcal{X}$ is a set of \emph{arguments}; 
$\mathcal{A}\!\subseteq\!\mathcal{X}\times\mathcal{X}$ is a relation \emph{of attack}; $\mathcal{S}\!\subseteq\!\mathcal{X}\times\mathcal{X}$ is a relation of \emph{support}; and
$\tau\!:\!\mathcal{X}\!\rightarrow\![0,1]$ is a total function, with $\tau(a)$ the \emph{base score} of $a\!\in\!\mathcal{X}$.


\begin{example}
\label{ex:qbaf}
Figure~\ref{fig:qbaf} 
depicts as a graph the QBAF $\langle\mathcal{X},\mathcal{A},\mathcal{S},\tau\rangle$ where 
    $\mathcal{X}=\{a,b,c\}$, $\mathcal{S}=\{(a, b)\}$, and $\mathcal{A}=\{(a, c)\}$ with $\tau(a)=0.5$, $\tau(b)=0.1$, and $\tau(c)=0.7$.
    \begin{figure}[htp!]
    \centering
    \includegraphics[width=.7\linewidth]{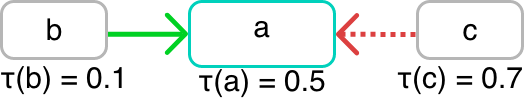}
    \caption{The QBAF 
    in Example~\ref{ex:qbaf} as a graph, with arguments represented as nodes, supports as green solid edges, and attacks as red dashed edges.
    }
    \label{fig:qbaf}
    \end{figure}
\end{example}


QBAFs are equipped with so-called \emph{gradual semantics}, assessing the (dialectical) \emph{strength} of arguments. 
 We use the \textbf{Discontinuity-Free Quantitative Argumentation Debate (DF-QuAD)} gradual semantics~\cite{Rago2016DFQuAD} to obtain the strength of the forecasters' opinions.
For a given QBAF $\langle\mathcal{X},\mathcal{A},\mathcal{S},\tau\rangle$, for any $x \in \mathcal{X}$ with $n\geq 0$ attackers with strengths $v_1, \ldots, v_n$, $m\geq 0$ supporters with strengths $v_1', \ldots, v_m'$ and $\tau(x)=v_{0}$, DF-QuAD computes $x$'s strength as
$\sigma(x)\!=\!\mathcal{C}(v_{0}, \mathcal{F}(v_{1}, \ldots, v_{n})$, $\mathcal{F}(v_{1}', \ldots, v_{m}'))$, where,
for any $w_{1}, \ldots, w_{k}$,  $\mathcal{F}(w_{1}, \ldots, w_{k})$ is $0$
if $k\!=\!0$  and $1 - \prod_{i=1}^k (|1-w_{i}|)$
otherwise and
$\mathcal{C}$ is defined as follows: for $v_{a}=\mathcal{F}(v_{1}, \ldots, v_{n})$ and $v_{s}=\mathcal{F}(v_{1}', \ldots, v_{m}')$,
if $v_a=v_s$ then $\mathcal{C}(v_{0}, v_{a}, v_{s})=v_0$; else if $v_a > v_{s}$ then $\mathcal{C}(v_{0}, v_{a}, v_{s})=v_{0} - (v_{0}\cdot|v_{s}-v_{a}|)$; otherwise $\mathcal{C}(v_{0}, v_{a}, v_{s})=v_{0} + ((1-v_{0})\cdot|v_{s}-v_{a}|)$.

\begin{example}
    Consider the QBAF 
    from Example~\ref{ex:qbaf}. Using DF-QuAD, $\sigma(b)=\tau(b)=0.1$ and $\sigma(c)=\tau(c)=0.7$ given that $b,c$ are neither attacked nor supported. 
    Then, 
    as $b$ is the single supporter and $c$ is the single attacker of $a$, and the latter is stronger than the former,
$\sigma(a)=\mathcal{C}(0.5, 0.7, 0.1)=0.5-(0.5\cdot 0.6)=0.2$
.
\end{example}

\textbf{ArguCast Frameworks}
We adopt 
a variant of 
the \textbf{\name} method from \cite{gorur2023argucast}, as follows.
An \emph{\name\ framework (\sname)} is a tuple $\langle\mathcal{X, R, U, V, P}\rangle$ such that:
$\mathcal{X=F\cup D}$ is a finite set of \emph{arguments} where $\mathcal{F}$ and $\mathcal{D}$ are disjoint; elements of $\mathcal{F}$  and $\mathcal{D}$ are referred to, respectively, as \emph{forecasting} and  \emph{non-forecasting arguments};
$\mathcal{R=A \cup S \subseteq D\times X}$, where $\mathcal{A}$ and $\mathcal{S}$ are disjoint relations 
of \emph{attack} and \emph{support}, respectively;
$\mathcal{U}$ is a finite set of \emph{forecasters};
$\mathcal{V:U\times D\rightarrow\{-,+,?\}}$ is a (partial) function where $\mathcal{V}(u, a)$ is the \emph{vote of forecaster} $u\in \mathcal{U}$ \emph{on (non-forecasting) argument} $a\in\mathcal{D}$;
$\mathcal{P:U\times F\rightarrow} [0,1]$ is a (partial) function where $\mathcal{P}(u, b)$ is the \emph{forecasting prediction by forecaster} $u\in\mathcal{U}$ \emph{on (forecasting) argument} $b\in\mathcal{F}$.
Our ACF frameworks are as in \cite{gorur2023argucast}, except for the introduction of forecaster votes $?$, which are absent in \cite{gorur2023argucast}.
We 
use these forecaster votes $?$ to indicate 
that the forecaster is unsure about arguments (e.g. because they are deemed irrelevant), distinguishing them from \emph{undefined} votes (i.e., for which $\mathcal{V}$ is undefined as the forecaster does not vote), indicating that the forecaster has no opinion on the arguments.    
Positive and negative votes are exactly as in \cite{gorur2023argucast}. Specifically, negative votes on arguments 
represent disagreement, indicating that the forecaster believes in the negation of those arguments.


\begin{example}
    \label{ex:acf}
    Consider 
    an \sname\ $\langle\mathcal{X, R, U, V, P}\rangle$ with the same arguments and relations as the QBAF 
    from Example~\ref{ex:qbaf}, 
    $\mathcal{U}=\{u\}$ with a single user disagreeing with $b$ (i.e., $\mathcal{V}(u, b)=-$) and agreeing with $c$ (i.e., $\mathcal{V}(u, c)=+$), and user $u$'s prediction on the forecasting argument 
    given by $\mathcal{P}(u, a)=0.85$.
    This is visualised in Figure~\ref{fig:acf}.
    \begin{figure}[htp!]
    \centering
    \includegraphics[width=.7\linewidth]{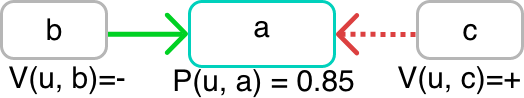}
    \caption{
    The ACF from Example~\ref{ex:acf} as a graph, with the forecasting argument represented as a cyan outlined node, the 
    other arguments 
    as gray outlined nodes, supports as green solid edges and attacks as red dashed edges.}
    \label{fig:acf}
    \end{figure}
\end{example}


For the remainder of the paper, unless specified otherwise, we assume as given an \sname\ $\langle\mathcal{X, R, U, V, P}\rangle$, with 
components as above.

\section{
Argumentative \Rationality\ of Forecasters}
\label{sec:methodology}
In this section, we define a novel 
way of identifying \irrational\ forecasters, 
based on 
their argumentative opinions and predictions. 
First, we define a novel notion of
forecasters 
as QBAFs drawn from \sname s
\FT{:}


\begin{definition}
\emph{A forecaster QBAF for} $u \in \mathcal{U}$ is a QBAF $\langle\mathcal{X}, \mathcal{A}_u, \mathcal{S}_u, \tau_u\rangle$ such that, for $a,b,c \in \mathcal{X}$
:
\begin{equation*}
\begin{split}
    \mathcal{S}_u=\{(b, c)\in \mathcal{R}|((b, c)\in\mathcal{S}\land(\mathcal{V}(u,b)=&\mathcal{V}(u,c) \\ \lor(\mathcal{V}(u,b)\neq-\land&\mathcal{V}(u,c)\neq-)) \\
    \lor((b,c)\in\mathcal{A}\land(
    \mathcal{V}(u,b)=-\land&\mathcal{V}(u,c)\neq-)
    ))\}
\end{split}
\end{equation*}
\begin{equation*}
\begin{split}
    \mathcal{A}_u=\{(b, c)\in \mathcal{R}|((b, c)\in\mathcal{A}\land(\mathcal{V}(u,b)=&\mathcal{V}(u,c) \\ \lor(\mathcal{V}(u,b)\neq-\land&\mathcal{V}(u,c)\neq-)) \\
    \lor((b,c)\in\mathcal{S}\land
    (\mathcal{V}(u,b)=-\land&\mathcal{V}(u,c)\neq-)
    ))\}
\end{split}
\end{equation*}

\begin{equation*}
\tau_u(a)
\begin{cases}
    \in [0,1]
    & \text{if} \ a\in\mathcal{F} \\
   = 0.5
     & \text{if} \ \mathcal{V}(u, a)\in\{+,-\}, \\
    =0 & \text{if} \ \mathcal{V}(u, a)\in\{?,\text{undefined}\}.
\end{cases}
\end{equation*}
\end{definition}


In a forecaster QBAF, we define $\mathcal{A}_u$ and $\mathcal{S}_u$ based on attack and support relations in ACFs: 
if a forecaster disagrees with an argument, we assume their opinion is the negation of that argument; thus, the relations to and from that argument are flipped, leading the argument to have the opposite effect 
than originally considered.
We keep the stance of a (child, parent) relation in $\mathcal{R}$ the same if the user has the same vote on both the child and parent argument or both of the arguments are not disagreed with. We change the stance of the relation if either one of the child or parent arguments is disagreed with and the other one is agreed with or is undecided (? or undefined).

The base score for the forecasting argument represents the confidence level of the forecaster on this argument prior to debating/voting.
We choose the [0,1] scale for base scores in-line with existing work in gradual semantics \cite{baroni2019gradual}. Higher assignments of this score mean there is stronger credibility for the forecasting argument, while lower assignments mean the forecasting argument is treated with more skepticism, and intermediate values provide a gradual scale between the two. Varying this base score allows us to control how much weight is given to the forecasting argument independently of the influence of supports and attacks.
Also, in a forecaster QBAF, an argument's base score is 0.5 
when the forecaster agrees or disagrees with the argument (we use 0.5 as it is the middle point between 0 and 1
).
Finally, in a forecaster QBAF, we set the base score to 0 for an argument that the user has not voted on, as we do not want the argument with no opinion to have an effect on the forecasting argument's strength.
The following example illustrates considerations for how the base score for forecasting arguments can be assigned.

\begin{example}
\label{ex:tau}
    Consider a forecasting argument $f$="X will win their next professional tennis match". We may assume a neutral assignment to the forecasting argument (e.g. $\tau(f)=0.5$) as we cannot make any assumptions on the initial confidence. 
    If we know X is an amateur tennis player, then we would expect a lower confidence level, hence a lower assignment to the forecasting argument (e.g. $\tau(f)=0.03$). 
    If instead we know X is the top-ranked player this season, then we would expect a higher confidence level, hence a higher assignment to the forecasting argument (e.g. $\tau(f)=0.7$).
    Thus, the base score reflects facts 
    known to the forecaster 
    prior to debate.
\end{example}

Our forecaster QBAFs differ from those underpinning the original ArguCast~\cite{gorur2023argucast}, specifically by satisfying properties of neutral forecasting (ensuring unbiased defaults when no opinions are expressed) and bias-based forecasting (ensuring opinions properly influence prediction strength), 
which
\cite{gorur2023argucast} fails to satisfy 
(see Appendix~\ref{app:proofs}).

In the remainder, unless specified otherwise, we  assume as given a forecaster QBAF $\langle\mathcal{X}, \mathcal{A}_u, \mathcal{S}_u, \tau_u\rangle$ for 
$u \in \mathcal{U}$ in the given \sname.



We introduce a notion to determine whether a forecaster is argumentatively \rational
. This notion constrains a forecasters' prediction 
to align with their internal reasoning. In particular, we assess the consistency between the forecasters' opinion (represented by the final strength of the forecasting argument within their QBAF), and their prediction, given two threshold values.
This is with the intention of 
filtering out users who do not satisfy this condition, to potentially gain more accurate forecasts.

\begin{definition}
\label{def:main}
    Given  functions $\xi_1:\mathcal{F}\rightarrow(0,1)$ and $\xi_2:\mathcal{F}\rightarrow(0,1)$\footnote{{We disallow threshold values of exactly 0 or 1 as then the \rationality\ condition would be too brittle in practice, since any minimal deviation would violate \rationality.}},
    forecaster $u$ is \emph{prediction \rational} 
    iff $\forall {f\in\mathcal{F}}$:
    \begin{itemize}
        \item if $\sigma(f) < \xi_1(f)$, then $\mathcal{P}(u, f) < \xi_2(f)$;
        \item if $\sigma(f) > \xi_1(f)$, then $\mathcal{P}(u, f) > \xi_2(f)$;
        \item if $\sigma(f) = \xi_1(f)$, then $\mathcal{P}(u, f) \in [\xi_2(f) - \epsilon, \xi_2(f) + \epsilon]$ for some small $\epsilon$.
    \end{itemize}
\end{definition}

This notion is more general than the one in \cite{gorur2023argucast} in that rather than fixing \rationality\ thresholds, we introduce two threshold functions $\xi_1$ and $\xi_2$, providing greater flexibility across forecasting contexts.


In the remainder, unless specified otherwise, $f$ will indicate a generic forecasting argument.

Intuitively, $\xi_1$ is a function which sets the midpoint of the strength of an argument which arbitrates whether the reasoning attacking or supporting it is more dominant. In the experiments, with arguments' base scores set up as in Definition \ref{def:main}, we use $\xi_1(f)=0.5$ throughout and leave to future work the study of other options.
Meanwhile, $\xi_2$ is a function that gives the initial or expected likelihood of the forecasting argument occurring, e.g. as in \cite{benjamin23HybridForecasting, tetlock2014forecasting_tournaments}.
In simple settings or when no prior knowledge is available, $\xi_2(f)$ may be set to 0.5, representing complete uncertainty, whereas higher or lower values may be set to the initial/expected probability, e.g. derived from expert judgment or prediction markets. Thus, $\xi_2(f)$ ensures that the forecasters' reasoning is grounded in an appropriate starting point.

\begin{example}
    Consider the forecasting argument from Example~\ref{ex:tau}, where the expected probability is a 50\% chance of X winning their next match, and thus we would assign a neutral value ($\xi_2(f)=0.5$). Now consider another forecasting argument `X will win the professional tennis competition', the probability is relatively low due to many competing players, so we may assign a lower value (e.g. $\xi_2(f)=0.2$).
    Finally, suppose we have a third forecasting argument `X will win a set in their next match',
    the expected probability is higher, and thus we would assign a higher value ($\xi_2(f)=0.8$).
\end{example}
Both threshold values are task-specific and thus need to be optimised for each task.
In the experiments, we use $\xi_2(f)=0.5$  or 
the initial probabilities from other judgmental forecasting platforms.

Finally, according to Definition~\ref{def:main}, a forecaster is considered \rational\ if their prediction aligns with their opinion on the forecasting argument. So, if they mostly support (or attack) the forecasting argument then their prediction should be higher (or lower) than the initial or expected probability. 

\section{In-Person Experiments}
\label{sec:inperson-experiment}

In this section we detail the findings from our first user study, which we undertook with members of the public attending the Great Exhibition Road Festival (GERF) in London in June 2024\footnote{\href{https://www.greatexhibitionroadfestival.co.uk/}{www.greatexhibitionroadfestival.co.uk}}.
The participants used  the ArguCast system
with pre-prepared questions on sports, politics and finance.
To assess the usefulness of \rationality, we then compared the outcomes of our variant of ArguCast (from now on referred to as ArguCast for simplicity) with those of baseline systems (seen as the ground truth here, considering the much larger basis of forecasters therein, compared to ArguCast).

\subsection{Research Question}


We evaluated the following Research Question (RQ):

\stepcounter{RQ}
{\bf RQ\theRQ} Do predictions from ArguCast align more with the predictions from other systems when 
\rationality\
is applied?
\\
We expected 
the null hypothesis 
(i.e., a negative answer to the question) 
to be rejected.

\subsection{Experimental Design}
To explore {\bf RQ\theRQ}, we asked participants to generate debates, express their opinions on the debates, and make predictions (in ArguCast) about three future events, along the following lines: ‘Will England win 
Euro 2024?’ (Euro), ‘Will the Labour party get a 50\% majority on the 4th of July 2024 UK general election?’ (Labour), and ‘Will Nvidia become the 
most profitable company in the world before the end of September 2024?’ (Nvidia).
Participants were given electronic devices such as tablets, laptops, and displays to use ArguCast.
The generated debates are illustrated in 
Appendix~\ref{app:in-person_debates},
and the forecaster QBAFs and predictions are available in the supplementary material.
Then, we applied our notion of \rationality\ 
with $\xi_1(f)=0.5$
and two options for $\xi_2$: $\xi_2(f)=0.5$, which is a neutral assignment, and $\xi_2(f)=\gamma$, where $\gamma$ is the initial prediction, which we obtained from prediction markets and Metaculus\footnote{Specifically, the initial predictions for the Euro and Labour questions
were taken from implied probabilities given by
\href{https://www.betfair.com/}{www.betfair.com}, and the baseline for the Nvidia question was taken from the prediction for this event on
\href{https://www.metaculus.com/questions/24806/nvidia-apple-market-cap-june-2024/}{www.metaculus.com} on the 19th of June 2024.}.
We then
computed the simple average of
both the raw predictions (
made by users) and \rational\ predictions (the predictions not filtered-out by \rationality), and
compared these predictions with the initial predictions by the baselines to assess the effect of the different values of $\xi_2$.

\subsection{Participant Recruitment and Ethics Statement}
We recruited 169 anonymous English-speaking participants at 
the GERF.\footnote{Some participants focused only on some of the questions, see Table~\ref{tab:gerf_experiment}.}
The data collection was anonymous as we created anonymous accounts for the participants on ArguCast. After reading the participant information sheet, participants who used the ArguCast system gave implied consent. Ethics approval was obtained from our institution prior to conducting the experiments.

\subsection{Results}

The results 
are shown in Table~\ref{tab:gerf_experiment}.%
\footnote{
Due to the small sample size and exploratory design, we did not perform 
statistical testing
.}
For the Euro question, the raw average prediction (38.06) was far from the initial prediction (20). Applying \rationality\ with $\xi_2(f)\!=\!0.5$ slightly moved the prediction further from the initial prediction, increasing it to (39.03). In contrast, \rationality\ with $\xi_2(f)\!=\!\gamma$ brought the prediction down to 34.89, closer to the initial prediction. These results were achieved by filtering out 52 and 56 participants for $\xi_2(f)\!=\!0.5$ and $\xi_2(f)\!=\!\gamma$, respectively.
For the Labour question, we see a similar trend where the raw average prediction (60.69) was far from the initial prediction (95). Applying \rationality\ improved the results, giving 61.40 with $\xi_2(f)\!=\!0.5$ and 63.33 with $\xi_2(f)\!=\!\gamma$. It should be noted that achieving these corrections required filtering out more than half of the participants: out of 54, 29 for $\xi_2(f)\!=\!0.5$ and 48 for $\xi_2(f)\!=\!\gamma$, indicating that a substantial proportion of \irrational\ responses diverged from the 
initial prediction.
For the Nvidia question, the raw average prediction (59.84) was lower than the initial prediction (80). Applying \rationality\ improved on the raw average predictions, 65.10 with $\xi_2(f)\!=\!0.5$ and 77.00 with $\xi_2(f)\!=\!\gamma$. Again we filtered out more than half of the participants, 18 for $\xi_2(f)\!=\!0.5$ and 6 for $\xi_2(f)\!=\!\gamma$.


\begin{table}[htp!]
\centering
\setlength{\tabcolsep}{0.5em}
\small
\begin{tabular}{ccccc}
\toprule
& \multirow{2}{*}{Initial ($\gamma$)} & \multirow{2}{*}{Raw Avg.} & \multicolumn{2}{c}{\Rational} \\
\cmidrule(lr){4-5}
& & & $\xi_2(f)=0.5$ & $\xi_2(f)=\gamma$
\\ \midrule
Euro & 
20 &
38.06 ($_{N=83}$) &
39.03 ($_{N=31}$) &
34.89 ($_{N=27}$)
\\
Labour  &
95 &
60.69 ($_{N=54}$) &
61.40 ($_{N=25}$) &
63.33 ($_{N=6}$)
\\
Nvidia  & 
80 &
59.84 ($_{N=38}$) &
65.10 ($_{N=20}$) &
77.00 ($_{N=6}$)
\\
\bottomrule
\end{tabular}
\caption{Predictions for the three forecasting questions comparing the initial 
prediction (by prediction markets or Metaculus), with the raw average of participant predictions on ArguCast, and the \rational\ predictions with 
$\xi_2(f)\!=\!0.5$ 
or $\xi_2(f)\!=\!\gamma$
.
$N$ is the number of participating users.
}
    \label{tab:gerf_experiment}
\end{table}

These results indicate that 
filtering out \irrational\ predictions can improve aggregate forecasts. 
Overall, across all three questions, applying \rationality\ with $\xi_2(f)=\gamma$ to \name\ consistently 
moved predictions closer to the initial predictions. However, when \rationality\ with $\xi_2(f)=0.5$ was applied the prediction moved further from the initial prediction for the Euro question.
This suggests that using a neutral \rationality\ threshold can still allow some of the incorrect predictions to be seen as \rational, thus pointing to the importance of a customisable notion of \rationality.
The results for the Labour and Nvidia questions further support this interpretation, as applying \rationality\ with $\xi_2(f)=\gamma$ led to greater improvements over compared to applying rationality with $\xi_2(f)=0.5$. 
The consistency in these improvements is valuable, particularly given the complexity and noise inherent in human judgment data \cite{kahneman2011thinking}.
Overall, we can reject the null hypothesis for {\bf RQ\theRQ},
since applying \rationality\ (for appropriate choices of the parameters improves forecasting accuracy. 

\section{LLM Experiments}
\label{sec:empirical_exp}

In this section, we present the findings of an empirical experiment conducted using LLMs on forecasting questions from GJOpen (a public judgmental forecasting platform \footnote{\href{https://www.gjopen.com/}{www.gjopen.com}}).

\subsection{Research Question}
Through these experiments, we address the following RQ:

\stepcounter{RQ}
{\bf RQ\theRQ} Does accuracy improve when \rationality\ is applied?

\noindent
We expected the null hypothesis 
(i.e., a negative answer to the question) 
to be rejected.

\subsection{Experimental Setup}
\subsubsection{Dataset}
We scraped 595 questions from GJOpen
. Some of these questions included both binary and multiple-choice answers, yielding 2923 answers in total.
Each question-answer pair was rephrased as a forecasting argument. For binary questions, we converted the pair into affirmative (or non-affirmative) statements of the forecasting question. For multiple-choice questions, we combined the answers with the question to form a forecasting argument. We generated these rephrasings using the Mistral-7B-Instruct-v0.3 LLM~\cite{jiang2023mistral}. All arguments were then manually reviewed, and any errors in the generated forecasting arguments were corrected.
The questions and the generated claims are available in the supplementary material.

\subsubsection{\name\ with ArgLLMs}

\begin{figure}
    \centering
    \includegraphics[width=\linewidth]{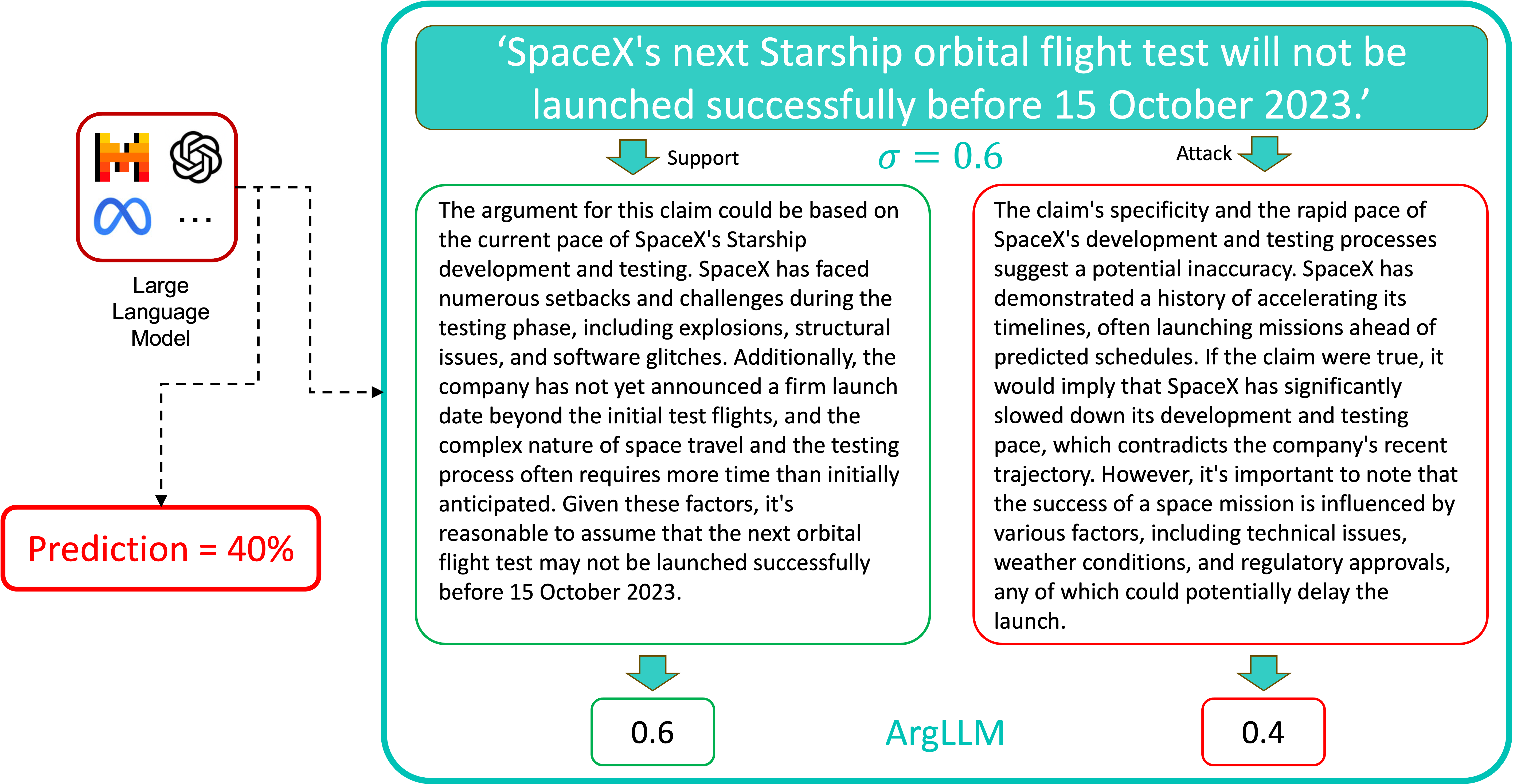}
    \caption{{An example of ArgLLM (with  Mistral-7B-Instruct-v0.3 as the underpinning LLM) applied to the forecasting argument $f$ 
    at the top, 
    giving a QBAF with a supporting and an attacking argument, with base scores 0.6 and 0.4 respectively, leading to a strength 
    $\sigma(f)\!=\!0.6$ with DF-QuAD (given 
    $\tau(f)\!=\!0.5$%
    ).} 
    The 
    {resulting forecaster QBAF leads to an  \irrational\ forecast, as the 
    40\% prediction is lower than $\xi_2(f)\!=\!0.5$, 
    but $\sigma(f) \!>\! \xi_1(f)\!=\!0.5$.
    }
    }
\label{fig:argllm_incoherent_incorrect}
\end{figure}

In order to populate \name, we used
LLMs to get predictions and 
ArgLLMs, 
leveraging LLMs, to construct QBAFs (we used the same prompts and hyperparameters as in \cite{freedman2024argllms}):
we prompted the LLMs to generate supporting and attacking arguments; for each generated argument, we prompted the LLMs again to obtain an uncertainty score (acting as the base score). 
We investigated two variations of ArgLLMs: generating arguments at breadth $1,1$ (only generating a supporter and an attacker for each forecasting argument), and breadth $n,k$ (generating 
$n$ supporting and $k$ attacking arguments for each forecasting argument). 
We set the base score for each forecasting argument 
to 0.5
.
Figure~\ref{fig:argllm_incoherent_incorrect} shows an example (at breadth $1,1$).  

After obtaining the predictions for the forecasting arguments
and the 
QBAFs, we apply \rationality
.


\subsection{Results}
We compared the overall accuracy with the accuracy of the \rational\ predictions (i.e., if the \rational\ predictions have a higher accuracy, then \rationality\ constraints would have a positive impact).
As LLMs, we used the Mistral (Mistral-7B-Instruct-v0.3)~\cite{jiang2023mistral}, Mixtral (Mixtral-8x7B-Instruct-v0.1)~\cite{jiang2024mixtral}, Llama 3 (Meta-Llama-3-8B-Instruct)~\cite{2024llama3}, and GPT-4o (GPT-4o-mini)~\cite{gpt-4o-mini} models.
Figure~\ref{fig:argllm_incoherent_incorrect} illustrates an \irrational\ and incorrect prediction, which will be filtered out leading to a more accurate prediction.

Table~\ref{tab:my_label} reports the forecasting accuracy for each LLM and the forecasting accuracy in the set of \rational\ predictions for both breadth $1,1$ and breadth $n,k$. The results indicate that applying \rationality\ generally leads to improvements or, at least, preserves forecasting accuracy across all models.
Note that applying \rationality\ significantly reduces the total number of predictions, at breadth $n,k$ generally retaining more samples compared to breadth $n,k$. Most notably, Llama 3 retains only 44\% of samples at breadth $1,1$, while it preserves 71\% at breadth $n,k$. The most samples retained were with the Mistral model at breadth $n,k$, preserving 78\% of the samples.

\begin{table}[htp!]
    \centering
    \footnotesize
    \setlength{\tabcolsep}{0.3em}
    \begin{tabular}{ccccccc}
          & \multicolumn{2}{c}{Raw LLM}
          & \multicolumn{2}{c}{ArgLLM(B$_{1,1}$)} 
          & \multicolumn{2}{c}{ArgLLM(B$_{n,k}$)} \\
          \cmidrule(lr){2-3} \cmidrule(lr){4-5} \cmidrule(lr){6-7}
           &
          Total &
          Acc. &
          Total &
          Coherent Acc. &
          Total &
          Coherent Acc.
          \\ \midrule
         Mistral
         & 2923
         & 77\% 
         & 1621
         & 80\% 
         & 2280
         & {\bf 81\%} 
         \\
         Mixtral 
         & 2923
         & 75\% 
         & 1693
         & {\bf 78\%} 
         & 1668
         & 75\%
         \\
         Llama 3
         & 2923
         & 75\%
         & 1294
         & {\bf 82\%} 
         & 2064
         & 79\% 
         \\
         GPT-4o
         & 2923
         & 69\%
         & 1716
         & {\bf 80\%}
         & 1922
         & 73\%
    \end{tabular}
    \caption{Accuracy comparison between raw LLMs and ArgLLMs with breadth $1,1$ and breadth $n,k$
    . For raw LLMs, we show accuracy as (correct/total). For ArgLLMs we show the percentages of \rational\ and correct predictions within those of \rational\ predictions (\rational\ \& correct/total \rational). The bold percentage shows the best variant for the given model. The total columns show the number of samples retained for each model.}
    \label{tab:my_label}
\end{table}


The key finding is that accuracy consistently improves when \rationality\ is applied, particularly at breadth $1,1$.
GPT-4o shows the most dramatic improvement, with accuracy increasing from 69\% to 80\% at breadth $1,1$.
Llama 3 and Mixtral also achieve their best performance when \rationality\ is applied at breadth $1,1$, reaching 82\% and 78\% accuracy, respectively.
Interestingly, Mistral is the only model that achieved a higher accuracy at breadth $n,k$ compared to breadth $1,1$, which could suggest that already better performing models may require more arguments to be generated.

These results support a positive answer to {\bf RQ\theRQ}
. Notably, the improvements are most consistent 
at breadth $1,1$.
Although small, the consistency of these improvements 
is valuable, as in the case of {\bf RQ1}.

\section{Crowd-sourced Experiments}
\label{sec:crowd-experiment}

Given the positive answers to \textbf{RQ1} and \textbf{RQ2}, in this section we report on the findings from our last user study with participants drawn from a crowd-sourcing platform, which aimed to assess the prevalence of \rationality\ by users in the real world. 
The participants were shown a number of forecasting debates following the ArguCast method, of different complexity. We then assessed participants' conformance to our notion of \rationality.

\begin{figure}[htp!]
    \centering
    \includegraphics[width=0.6\linewidth]{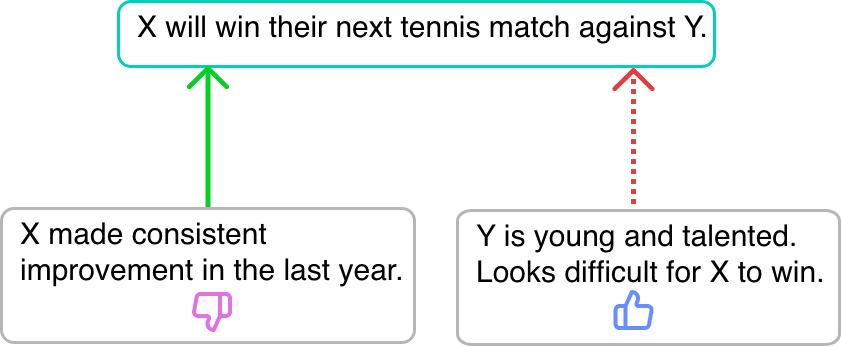}
    \caption{Example of a debate 
    shown to users. 
    This includes a forecasting argument (
    in cyan), a supporter thereof (
    in gray with a green solid arrow pointing towards the forecasting argument), an attacker of the forecasting argument (also 
    in gray but with a red dotted arrow pointing towards the forecasting argument), and the votes of a fictitious forecaster $u$ (where blue thumbs up represents $V(u, a)=+$ and pink thumbs down represents $V(u, a)=-$).}
    \label{fig:example_debate_variation}
    
\end{figure}

\subsection{Research Questions}
We aim to evaluate 
the following 
RQs:

\stepcounter{RQ}
\edef\theRQa{\theRQ}
{\bf RQ\theRQa} Does our notion of 
\rationality\ {\em align} with users' reasoning in the context of forecasting?

\stepcounter{RQ}
\edef\theRQb{\theRQ}
{\bf RQ\theRQb} Does our notion of  \rationality\ align {\em more} with users' reasoning when debates are not ``complex''?
\\
We anticipated, for {\bf RQ\theRQa}, the null hypothesis (stating that users' reasoning and our notion of \rationality\ disagree in the same manner) 
not to be rejected. However, for {\bf RQ\theRQb}, we expected the null hypothesis to be rejected (i.e., that {\bf RQ\theRQb} 
is answered affirmatively).


We focused on three ``complexity'' types, 
along three axes: vote, breadth, and depth
given below 
(for 
more details see Appendix~\ref{app:complexities}).

\begin{definition}
    A \emph{simple variant} is an \sname\ $\langle\mathcal{X}=\mathcal{F}\cup \mathcal{D}, \mathcal{R}=\mathcal{A}\cup \mathcal{S}, \mathcal{U}, \mathcal{V}, \mathcal{P}\rangle$ such that: 

    \hspace*{0.5cm}\(
        |\mathcal{F}|=1 \land |\mathcal{D}|=2 \land |\mathcal{A}|=1 \land |\mathcal{S}|=1 \land 
        (a,f) \in \mathcal{A} 
        \rightarrow\)
        
        \hspace*{0.5cm}\(\mathcal{V}(u, a)=+ \land 
        (s,f) \in \mathcal{S} 
        \rightarrow\mathcal{V}(u, s)=-.
    \)
\end{definition}

So, a \emph{simple variant} has three arguments: the forecasting argument, and a supporter and an attacker 
thereof; and the forecaster disagrees with the supporter and agrees with the attacker.

\begin{definition}
    A \emph{vote complex variant} is an \sname\ $\langle\mathcal{X}=\mathcal{F}\cup \mathcal{D}, \mathcal{R}=\mathcal{A}\cup \mathcal{S}, \mathcal{U}, \mathcal{V}, \mathcal{P}\rangle$ such that:
    
    \hspace*{0.2cm}\(
        (\exists_{b\in \mathcal{D}}\exists_{f\in\mathcal{F}}(b,f)\in\mathcal{A}\land\mathcal{V}(u,b)= -)\lor\)

        \hspace*{0.2cm}\(
        (\exists_{c\in \mathcal{D}}\exists_{d\in \mathcal{D}}\exists_{g\in\mathcal{F}}(c,g)\in\mathcal{A}\land(d,g)\in\mathcal{S}
        \land\mathcal{V}(u,c)=\mathcal{V}(u,d)).
        \)
\end{definition}
Thus, a \emph{vote complex variant} is where the forecaster disagrees with an argument that attacks the forecasting argument (giving a double negative), or if the forecaster has conflicting votes (e.g. if the forecaster agrees with both arguments that support and attack the forecasting argument). The votes become more complex because the idea of disagreeing with an argument in an attack relation and conflicting votes is harder for some forecasters to understand\footnote{We found this in a small pilot experiment with 12 participants.}.


    



\begin{definition}
    A \emph{breadth complex variant} is an \sname\ $\langle\mathcal{X}=\mathcal{F}\cup \mathcal{D}, \mathcal{R}=\mathcal{A}\cup \mathcal{S}, \mathcal{U}, \mathcal{V}, \mathcal{P}\rangle$, such that:
    
    \hspace*{1cm}\(
    \lvert\{a\in \mathcal{D}|\not\exists_{r\in \mathcal{R}}{(r=(a_1,a), a_1\in\mathcal{D})}\}\rvert=3
    \).
\end{definition}

A \emph{breadth complex variant} occurs when the forecaster adds 
a new supporter or attacker to the forecasting argument, i.e., the forecasting argument has three child arguments\footnote{\label{fn:new_arg}Note that since the forecaster adds the new argument, their vote on that argument is automatically set to agreement.}.

\begin{definition}
    A \emph{depth complex variant} is an \sname\ $\langle\mathcal{X}=\mathcal{F}\cup \mathcal{D}, \mathcal{R}=\mathcal{A}\cup \mathcal{S}, \mathcal{U}, \mathcal{V}, \mathcal{P}\rangle$ such that:
    
    \hspace*{1cm}
    \(
        |\{a\in \mathcal{D}|\exists_{r\in \mathcal{R}}{(r=(a,a_1), a_1\in\mathcal{X})}\}|=1
\).
\end{definition}

A \emph{depth complex variant} occurs when the forecaster adds a new argument to any of the leaf (i.e. unattacked and unsupported) nodes. This argument can either support or attack one of the child arguments (the supporter or attacker of the forecasting argument).\textsuperscript{\ref{fn:new_arg}} 


    

With these notions of complexity, we can refine {\bf RQ\theRQb} into:

{\bf RQ\theRQb.1} Does our notion of  \rationality\ align {\em more} with users' reasoning when debates are not \emph{vote complex}?

{\bf RQ\theRQb.2} Does our notion of  \rationality\ align {\em more} with users' reasoning when debates are not \emph{breadth complex}?

{\bf RQ\theRQb.3} Does our notion of  \rationality\ align {\em more} with users' reasoning when debates are not \emph{depth complex}?

\subsection{Experimental Design}

In order to answer the RQs we ran a user experiment on the online recruitment platform Prolific\footnote{Prolific (\url{www.prolific.com}) was chosen for its superior data quality compared to other similar platforms \cite{Per2021DataQO}.} using a  Qualtrics survey. 
In the survey, each user was assigned four variants of a debate, chosen randomly between two available debates for two different binary forecasting questions, with abstract entities to prevent bias (i.e., {\em Will X win their next tennis match against Y?} and {\it Will Party A win the next election against Party B?}), where each debate has 60 potential variants. The (variants of the) debates capture votes of a fictitious forecaster Alex and Alex’s prediction on the forecasting question. The full debates are shown in 
Appendix~\ref{app:debates}.
Figure~\ref{fig:example_debate_variation} gives a simple example thereof
.

The 
survey follows this design flow:
    {\bf (1)} Main Test (4 x 3 questions) -- here, we provide the users with four (variants of) debates, each containing Alex’s prediction, part of the debate with Alex’s votes, and three questions,  consisting of an attention check, whether the user agrees with Alex’s prediction, and their feedback on their reasoning (optional). This task addresses both research questions ({\bf RQ\theRQa\ and RQ\theRQb}). 
    {\bf (2)} Post-tasks (4 questions) -- the participants are asked questions regarding their experience with the questionnaire.

In this experiment, we set $\xi_1(f)=\xi_2(f)=0.5$.
We choose $\xi_2(f)=0.5$ due to the forecasting questions being fictitious and having a 50\% probability of occurring.
The participants were divided into two groups: 50\% of them were asked to assess simpler questions first and more complex questions 
later (named the Simple-Complex group) while the other 50\% of were asked to assess more complex questions first and simpler questions after 
(the Complex-Simple group). In both cases, users were randomly allocated (50\%, 50\%) which (of the two) forecasting questions to assess
. For the four (variants of) debates the users saw, Alex’s forecast was also allocated randomly (1/3, 1/3, 1/3) amongst $<50\%,=50\%, >50\%$.
In 
Appendix~\ref{app:experimental-design}
we further 
discuss the experimental design of this study.


For an expected significance of 0.05, a power of 0.8, and an effect size of 0.3, we needed 174 samples for {\bf RQ\theRQa} (where a $\chi^2$ test is used) and 90 for each {\bf RQ\theRQb.i}. We ended up recruiting 100 participants as this yielded 400 samples for {\bf RQ\theRQa} and 200 samples for each {\bf RQ\theRQb.i}, or 360 for {\bf RQ\theRQa} and 180 for each {\bf RQ\theRQb.i} accounting for a 10\% dropout, which was sufficient.

\subsection{Participant Recruitment and Ethics Statement}
We recruited 100 anonymous English-speaking participants aged 18+ from Prolific, paying them an average of £4.50 for their time, based on the platform's standard reimbursement rate of £26.82 per hour. The data collection was anonymous
. Participants were asked to provide informed consent after reading the participant information sheet, and they could withdraw at any time before completing the survey by simply closing the questionnaire window, with their data permanently deleted. Participants were informed that once they completed the survey, they could no longer withdraw, and we would retain their data. Ethics approval 
was obtained from our institution prior to conducting the experiments.

\subsection{Results}
\label{sec:results}
Out of 100 participants, only 46 passed all attention checks
and were included in our study. For {\bf RQ\theRQa} we analysed 184 samples using McNemar's test \cite{McNemar1947} used for determining whether two classification models disagree in the same way. The null hypothesis was that the two marginal probabilities for each outcome (from our notion and users) were identical
(i.e., H0: $12=76$, H1: $12\neq76$ from the contingency Table~\ref{tab:contingency}). The chi-squared test (commonly used in conjunction with McNemar’s) with one degree of freedom yielded $p<.00001$. Given the result is significant at $p<0.05$, we rejected H0 in favour of H1, indicating that the users' opinion of \rationality\ does not align with our notion
.

\begin{table}[htp!]
    \centering
    \begin{tabular}{cccc}
    & & \multicolumn{2}{c}{\bf User \Rational?} \\
    \cmidrule(lr){3-4}
        & & Yes & No  \\
        \multirow{2}{*}{\bf \makecell{Model \\ \Rational?}}& \multicolumn{1}{|c}{Yes} & 44 & 12 \\
        & \multicolumn{1}{|c}{No} & 76 & 52 \\
    \end{tabular}
    \caption{Contingency table for McNemar's test. Columns/rows represent the number of debate variants the users/our notion of \rationality\ (respectively) found \rational\ 
    for the same debate
    variants.}
    \label{tab:contingency}
\end{table}

\begin{table}[htp!]
    \centering
    \small
    \setlength{\tabcolsep}{0.79em}
    \begin{tabular}{ccccccccc}
    \cline{2-9}
    & s & v & b & d & d/b & v/d & v/b & v/d/b \\
    \hline
         {\bf Aligned} & 20 & 7 & 7 & 9 & 9 & 8 & 9 & 27 \\
         {\bf $\neg$Aligned} & 26 & 9 & 6 & 8 & 9 & 3 & 8 & 19 \\ 
    \hline
    \end{tabular}
    \caption{Number of (not) aligned and 
    pairs (alignment between our notion of \rationality\ and the participants in the experiment) for the debate variations: \underline{s}imple,
    \underline{b}readth complex, 
    \underline{d}epth complex and
    \underline{v}ote complex. 
    }
    \label{tab:variant_data}
\end{table}

Table~\ref{tab:complexity} summarises the results for {\bf RQ\theRQb.1-RQ\theRQb.3}. For each complexity type, we conducted a two-sample one sided t-test to determine if complex debate variants aligned more with our notion of \rationality\ compared to the simpler variants. Our null hypothesis for all {\bf RQ\theRQb.1-RQ\theRQb.3} was H0: $\mu_{\text{complex}}\leq\mu_{\neg \text{complex}}$ against the alternative H1: $\mu_{\text{complex}}>\mu_{\neg \text{complex}}$. All three complexity types were statistically significant at $p<0.05$. Vote complexity yielded a p-value of 0.009, breadth complexity a p-value of 0.04, and depth complexity the most significant effect with a p-value of 0.002. Based on these results, we can reject all three null hypotheses
.

\begin{table}[htp!]
    \centering
    \footnotesize
    \setlength{\tabcolsep}{0.45em}
    \begin{tabular}{cccccc}
         \!Complexity\! & 
         \!$\mu_\text{complex}$\! & 
         \!$\mu_{\neg \text{complex}}$\! & 
         \!$SD_\text{complex}$\! & 
         \!$SD_{\neg \text{complex}}$\! & 
         \!p-value\! \\
         \midrule
         Vote & 0.57 & 0.48 & 0.245 & 0.25 & 0.009 \\
         Breadth & 0.55 & 0.49 & 0.247 & 0.25 & 0.04 \\
         Depth & 0.58 & 0.47 & 0.24 & 0.25 & 0.002 \\
         \bottomrule
    \end{tabular}
    \caption{Mean and standard deviation of aligned pairs (between our notion of \rationality\ and the participants from the user experiment) for each complex and not-complex debate variations. These values are derived from Table~\ref{tab:variant_data}. Then, it reports on the p-value derived from the two-sample one sided t-test.}
    \label{tab:complexity}
\end{table}

Rejecting {\bf RQ\theRQa} suggests that the notion of \rationality\ with $\xi_1(f)=\xi_2(f)=0.5$ does not accurately reflect the intuition of what users perceive as \rational. This further implies the need for a \rationality\ constraint, as users are not naturally \rational\ in their judgment.
Additionally, since we have rejected {\bf RQ\theRQb.1, RQ\theRQb.2, and RQ\theRQb.3} and these were refinements of {\bf RQ\theRQb}, we can also reject {\bf RQ\theRQb}. This means that our notion of \rationality\ aligns more with users' reasoning when debates are complex. This could suggest that our notion of \rationality\ is more intuitive when more information is available to the users. Furthermore, this indicates that with complex debate variations, the users spend more cognitive effort with complex debate variations, which helps them counteract their biases. We have left the qualitative error analysis of forecaster biases to future work.


\section{
Conclusion}
\label{sec:conclusion}

In this paper,  
we introduced a novel notion of argumentative \rationality\ for judgmental forecasting, formally defined in terms of alignment between forecasters' internal argumentative reasoning and their predictions.
Then, through crowd-sourced and in-person user studies, as well as
 experiments with LLMs, we systematically evaluated the usefulness of the \rationality\ property.

First, in an in-person user study with 
ArguCast, we found that filtering out \irrational\ forecasts improved the aggregated predictions, moving them closer to expert baselines. Second, in the experiments with LLMs, we applied \rationality\ to LLM-generated forecasts and ArgLLM-generated argumentative reasoning and demonstrated consistent improvements in accuracy across different models.
These results from the in-person and LLM experiments support the practical need for integrating argumentative \rationality\ into judgmental forecasting systems (
ArguCast, GJOpen, Metaculus).
Finally, in the crowd-sourced experiments, we found that the users' opinions may not align with (our notion of) \rationality.
This 
suggests that 
some form of \rationality\ constraints are necessary.
%
Additionally, we observe that the users' opinions aligned more with our notion of \rationality\ when debates  were more complex. We speculate that this is 
due to our notion being more intuitive when more information is available
.
The results also indicated that as users spend more cognitive effort with complex debate
s (which implies that they reason more), it helps them counteract their biases. 
This may have 
implications for works considering argumentative models of 
reasoning, e.g. \cite{Rago_23}.

For future work, we plan to run a similar study to the crowd-sourced study with argumentation experts to determine whether our notion of \rationality\ is 
intuitive to them.
Additionally, we aim to extend our notion of \rationality\ to provide a degree of \rationality\ rather than a binary outcome. This would allow us to weight predictions, giving more influence to those that are more \rational, thereby enhancing the impact on the group forecast.
We also plan to explore relationships between our approach and probabilistic argumentation~\cite{toni2023forecasting}.
Finally, we would like to use  FORECAST2023 \cite{GorskaL2024forecast2023} and/or the hybrid-forecasting competition \cite{benjamin23HybridForecasting} dataset to populate an argumentation-based forecasting system (e.g. ArguCast) and study whether the notion of \rationality\ improves forecasting accuracy, as was the case for 
the notion of \rationality\ in \cite{irwin2022forecasting}.

\section*{Acknowledgments}
This research was partially funded by the ERC under the EU’s Horizon 2020 research and innovation programme (grant agreement no. 101020934, ADIX), by J.P. Morgan and by the Royal Academy of Engineering, UK, under the Research Chairs and Senior Research Fellowships scheme (grant agreement no. RCSRF2021\textbackslash 11\textbackslash 45).


\bibliographystyle{named}
\bibliography{bibliography}

\clearpage
\appendix

\section*{Supplementary Material }
\section{Desirable Properties of Forecaster QBAFs}
\label{app:proofs}


We introduce two properties that provide sanity checks for the forecaster QBAFs, which is the underlying framework for our notion of \rationality.

\begin{property}(Neutral Forecasting) 
$\forall f\in\mathcal{F}$:
\begin{align*}
    \forall_{a\in\mathcal{D}}{\mathcal{V}(u, x)\in\{?,\text{\small undefined}\}} 
\rightarrow\sigma(f)= 0.5 
\end{align*}
\end{property}

This property ensures that if a forecaster has not voted on any argument then the strength of the forecasting argument $f$ should be $0.5$, which is the base score of the forecasting argument. So, based on Definition~\ref{def:main}, if the forecaster has not expressed their opinion then their prediction should be close to $\xi_2(f)$. This will force forecasters to vote (i.e., show their reasoning) if they want to make a prediction on $f$ further away from $\xi_2(f)$.



\begin{proposition}
    Forecaster QBAFs {\em satisfy} the Neutral Forecasting property.
\end{proposition}

\begin{proof}
    We need to show that in forecaster QBAFs, if  $\mathcal{V}(a)=?$ $\forall a\in\mathcal{X}$ then $\forall f \in \mathcal{F}$ $\sigma(f)=0.5$ holds.
    It can be seen, by Definition 1, that if $a \not\in \mathcal{F}$ and $\mathcal{V}_u(a)=?$ then $\tau(a) = 0$, and if $a \in \mathcal{F}$ then $\tau(a) = 0.5$.
    Since the forecaster QBAF is a tree, we first consider some $a$ which is a leaf, i.e., $\nexists b \in \mathcal{X}$ such that $(b,a) \in \mathcal{A} \cup \mathcal{S}$, by the definition of DF-QuAD, we know that $\sigma(a) = \tau(a)$. Thus, for any $f \in \mathcal{F}$ which is a leaf, the property holds.
    If $\exists f \in \mathcal{F}$ which are not leaves, we calculate the strengths of the arguments which are only attacked or supported by leaves, and as shown by the fact that DF-QuAD satisfies \cite{amgoud2018evaluation} {\em Principle 6 (neutrality)}, as well as the fact that, by Definition 1, $(\mathcal{F} \times \mathcal{X}) \cap (\mathcal{A} \cup \mathcal{S}) = \emptyset$, we know that for these arguments also, $\sigma(a) = \tau(a)$ holds by the definition of DF-QuAD. We then continue to propagate up the tree in the same way, until $\forall f \in \mathcal{F}$, it is shown that $\sigma(f) = 0.5$, and thus the property holds.
    
\end{proof}

Note that this is not the case from existing work \cite{gorur2023argucast}, consider a single attacker $(a,f) \in \mathcal{A}^u$ of the forecasting argument with $V_u(a)=?$, then the strength for the forecasting argument would be $\sigma(f)=0.25$. This is a counterexample that shows the Neutral Forecasting property is not satisfied for their notion of \rationality.



\begin{property}(Bias-based Forecasting) For $f\in\mathcal{F} 
$:
\begin{align*}
    \exists_{(a,f)\in\mathcal{A}_u}
    \tau_u(a)=1\land  \forall_{x\in\mathcal{D} \setminus \{ a \}}\mathcal{V}(u, x)\in\{?,\text{\small undefined}\} \\ 
    \rightarrow\sigma(f)<0.5 
    \ \text{and} \\
    \exists_{(s,f)\in\mathcal{S}_u}
    \tau_u(s)=1\land\forall_{x\in\mathcal{D} \setminus \{ s \}}\mathcal{V}(u, x)\in\{?,\text{\small undefined}\} \\ 
    \rightarrow\sigma(f)>0.5 
\end{align*}
\end{property}

If the forecaster agrees with only the attackers (or supporters), the final strength $\sigma(f)$ should be less (or more) than $0.5$. This differs from the notion in \cite{gorur2023argucast}, where when there were many (more than 50) supporters (or attackers) which the forecaster had no opinion and the only argument with which the forecaster agrees is an attacker (or supporter), then the strength would be exactly $\sigma(f)=0.5$, which is the base score. This would identify forecasters who exhibit bias, as their prediction would deviate from their opinions.

\begin{proposition}
    Forecaster QBAFs {\em satisfy} the Bias-based Forecasting property.
\end{proposition}
\begin{proof}
    Consider the first case, in which we need to show that $\sigma(f)<0.5$.
    The proof showing that $\forall x \in \mathcal{X} \setminus \{ f \}$, $\sigma(x) = \tau(x)$ is the same as that for Proposition 1. We thus know that $\sigma_\mathcal{S}^u(f)=0$ and then, since $\sigma(a)=\tau(a)=1$, $\sigma_\mathcal{A}^u(f) = 1$. Thus, it can be seen by the definition of DF-QuAD that $\sigma(f)<\tau(f) = 0.5$ if $\sigma_\mathcal{A}^u(f)>\sigma_\mathcal{S}^u(f)$ (this was shown in \cite{potyka2021interpreting} {\em (strict monotonicity)} excluding when $\tau(f)=0$ and $\tau(f)=1$, which is not the case here).

    The proof for the second case is analogous, therefore, the Bias-based Forecasting property is satisfied for forecaster QBAFs as both conditions of the property are satisfied.

\end{proof}



So, our notion of \rationality\ has advantages over that in \cite{gorur2023argucast}, which satisfies neither of these properties. We leave a full theoretical analysis of our method's properties to future work, given the focus on the user studies in this paper.


\section{Debates Generated in the In-Person Experiments}
\label{app:in-person_debates}
\begin{figure}[htp!]
    \centering
    \includegraphics[width=\linewidth]{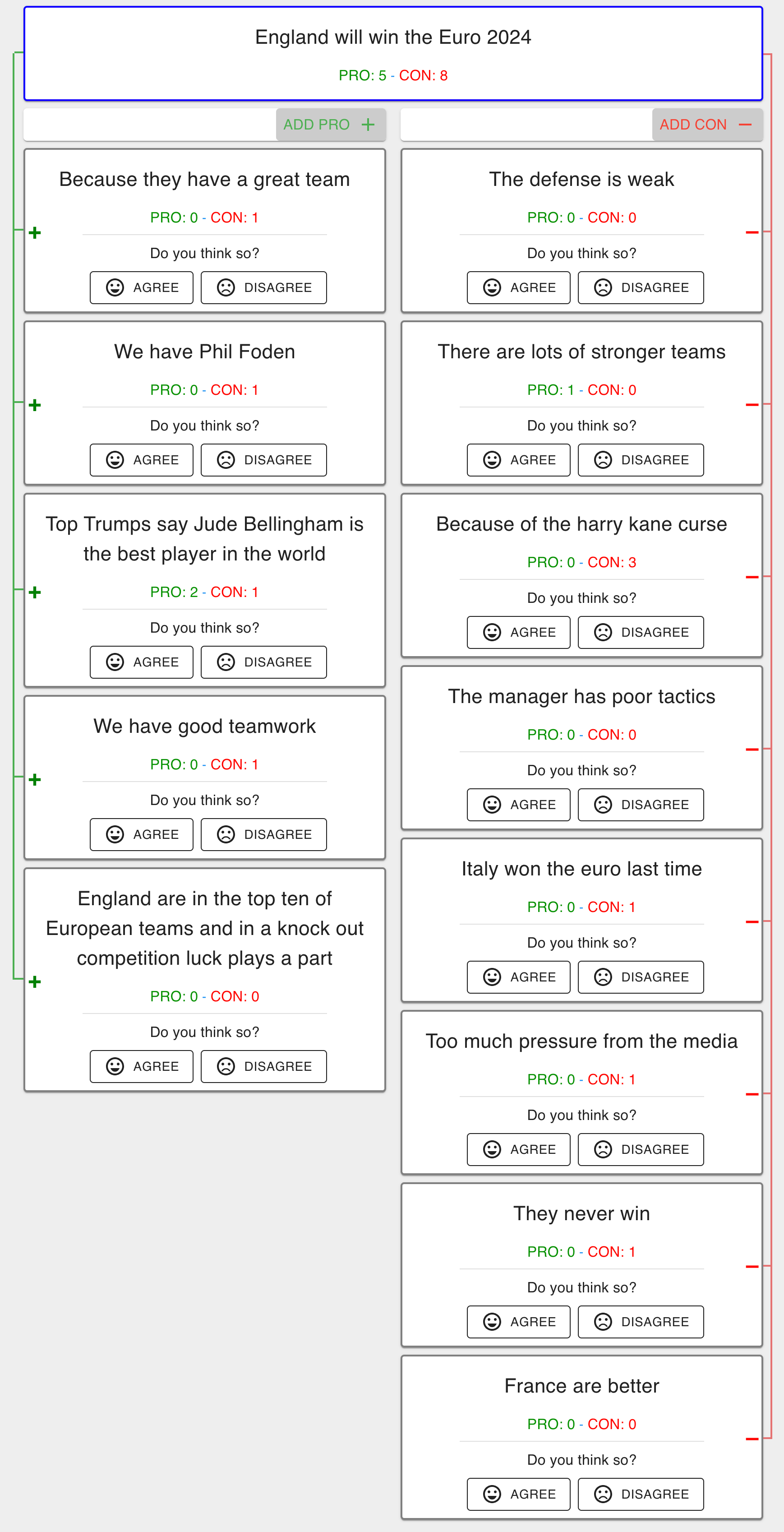}
    \caption{Supporting and attacking arguments generated for the Euro question by the participants of the in-person experiment.}
    \label{fig:euro}
\end{figure}

\begin{figure}[htp!]
    \centering
    \includegraphics[width=\linewidth]{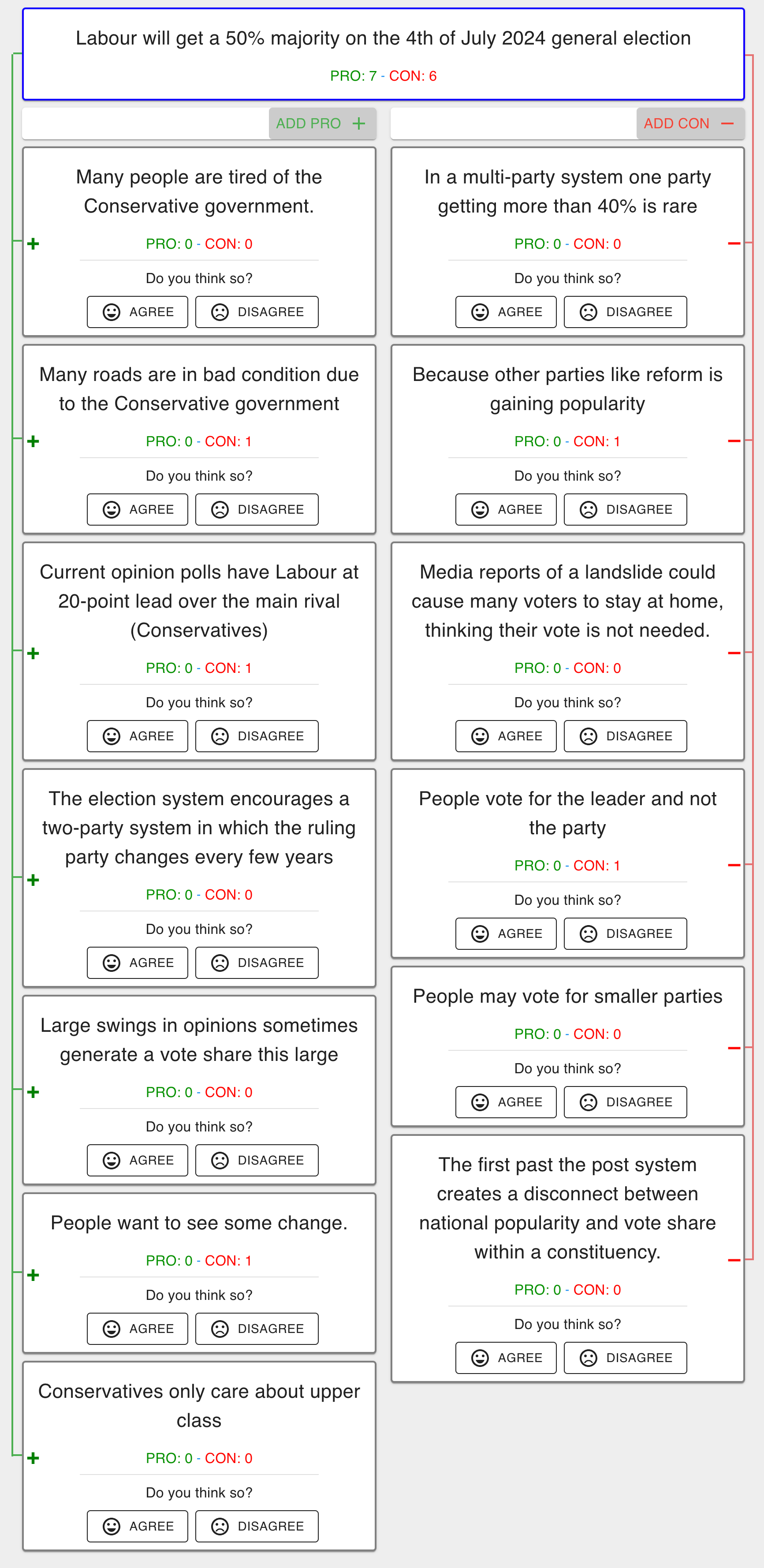}
    \caption{Supporting and attacking arguments generated for the Labour question by the participants of the in-person experiment.}
    \label{fig:labour}
\end{figure}

\begin{figure}[htp!]
    \centering
    \includegraphics[width=\linewidth]{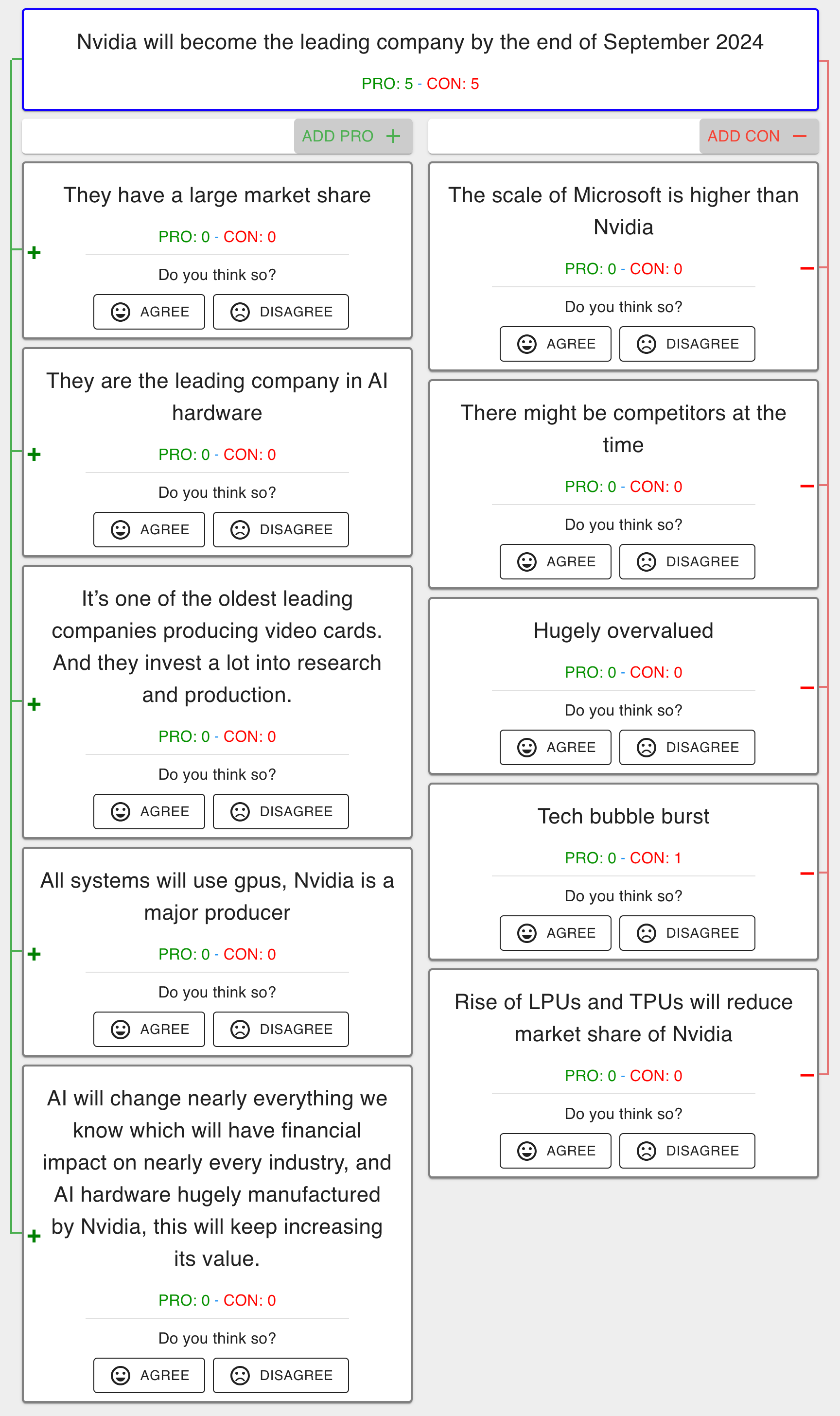}
    \caption{Supporting and attacking arguments generated for the Nvidia question by the participants of the in-person experiment.}
    \label{fig:nvidia}
\end{figure}

The debates generated for the three questions in the in-person experiment are given in Figure~\ref{fig:euro}, \ref{fig:labour}, and \ref{fig:nvidia}.

\section{Full Debates on Forecasting Questions}
\label{app:debates}
\begin{figure*}[htp!]
    \centering
    \includegraphics[width=\linewidth]{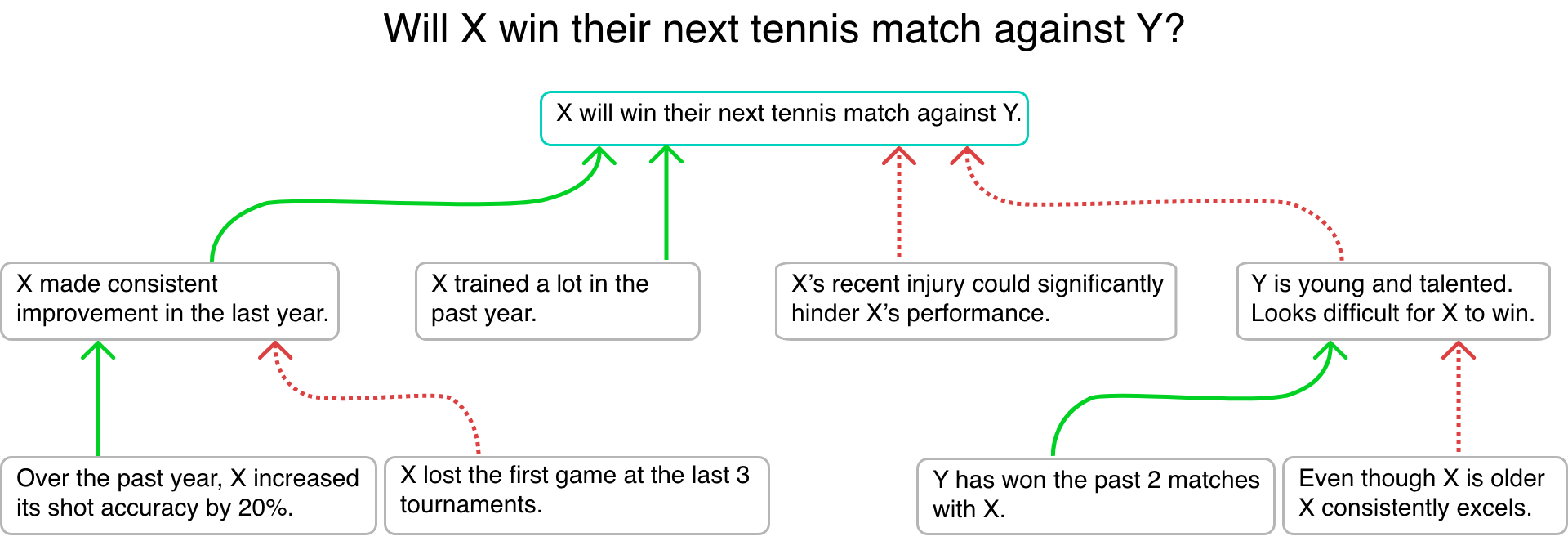}
    \caption{Full debate for the forecasting question `Will X win their next tennis match against Y?'.}
    \label{fig:debate1}
    
\end{figure*}

\begin{figure*}[htp!]
    \centering
    \includegraphics[width=\linewidth]{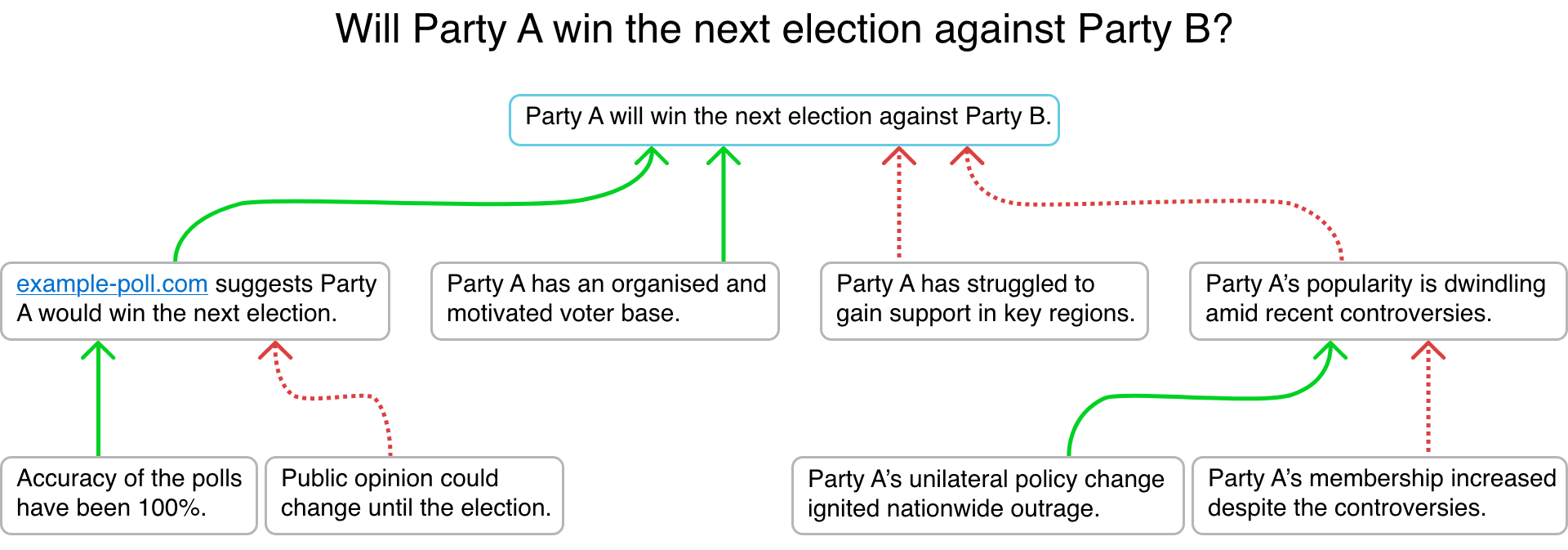}
    \caption{Full debate for the forecasting question `Will Party A win the next election against Party B?'.}
    \label{fig:debate2}
    
\end{figure*}

The full debates used in our experiments are given in Figure~\ref{fig:debate1} and Figure~\ref{fig:debate2}.

\newpage
\section{Abstract Structure of Complexities}
\label{app:complexities}
\begin{figure}[htp!]
    \centering
    \includegraphics[width=0.3\linewidth]{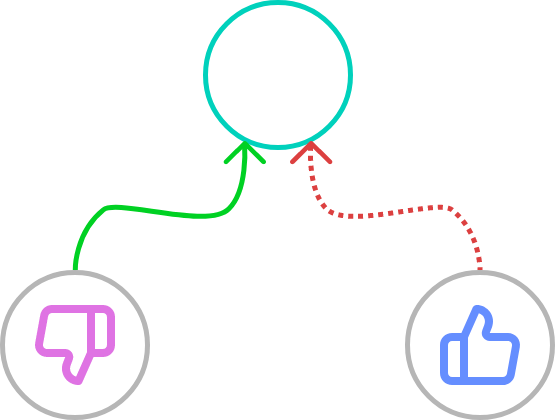}
    \caption{Abstract structure of the simple debate variant.}
    \label{fig:simple}
    
\end{figure}

\begin{figure}[htp!]
    \centering
    \includegraphics[width=0.8\linewidth]{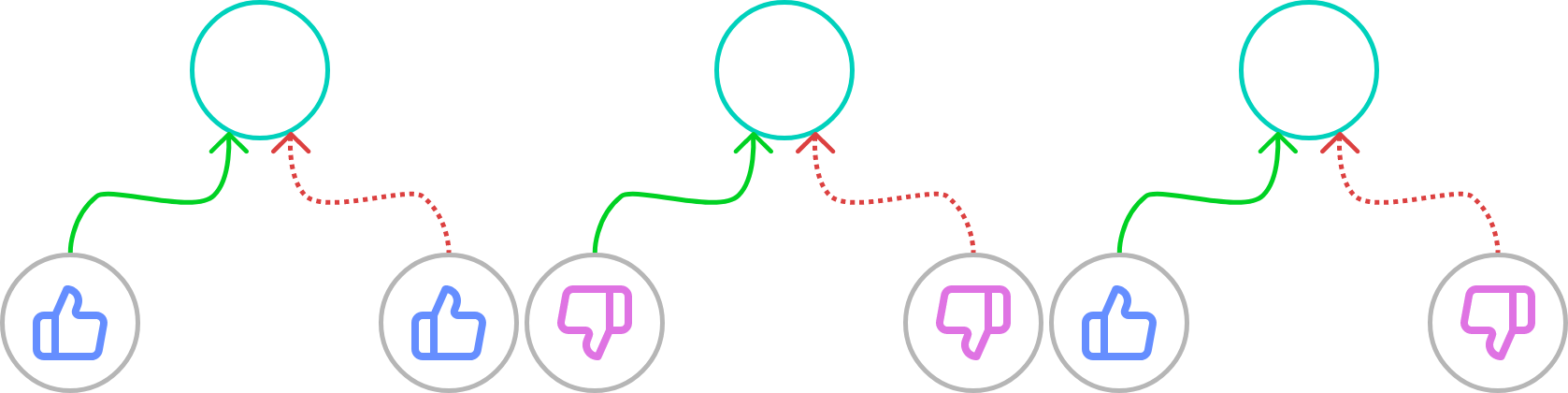}
    \caption{Abstract structure of the vote complex debate variant.}
    \label{fig:vote_complexity}
    
\end{figure}

\begin{figure}[htp!]
    \centering
    \includegraphics[width=0.5\linewidth]{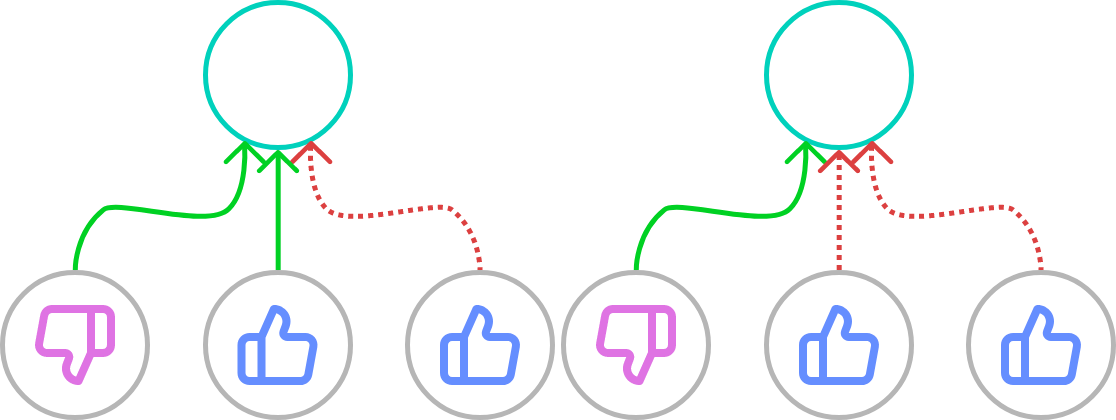}
    \caption{Abstract structure of the breadth complex debate variant.}
    \label{fig:breadth_complexity}
    
\end{figure}

\begin{figure}[htp!]
    \centering
    \includegraphics[width=0.5\linewidth]{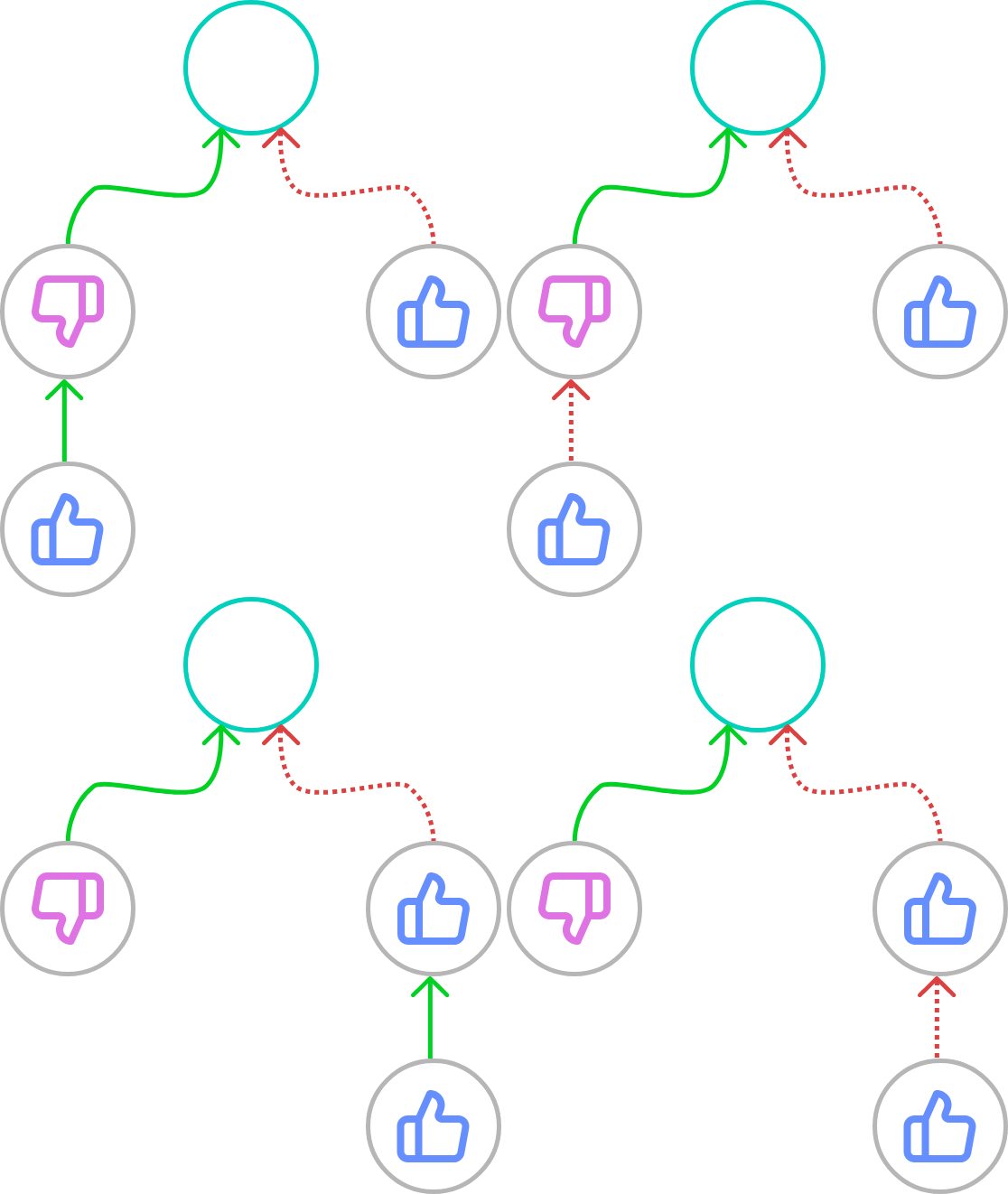}
    \caption{Abstract structure of the depth complex debate variant.}
    \label{fig:depth_complexity}
    
\end{figure}

\begin{figure}[htp!]
    \centering
    \includegraphics[width=0.5\linewidth]{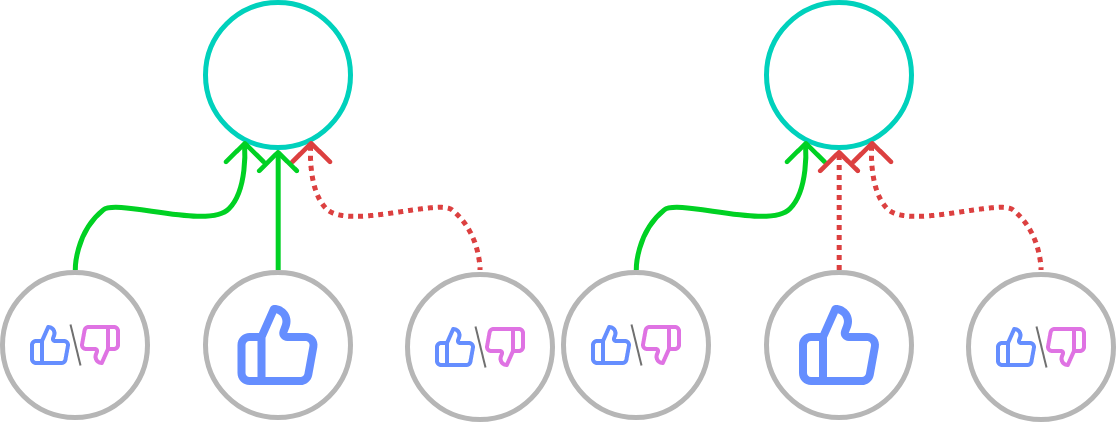}
    \caption{Abstract structure of the vote/breadth complex debate variant.}
    \label{fig:vote-breadth_complexity}
    
\end{figure}

\begin{figure}[htp!]
    \centering
    \includegraphics[width=0.5\linewidth]{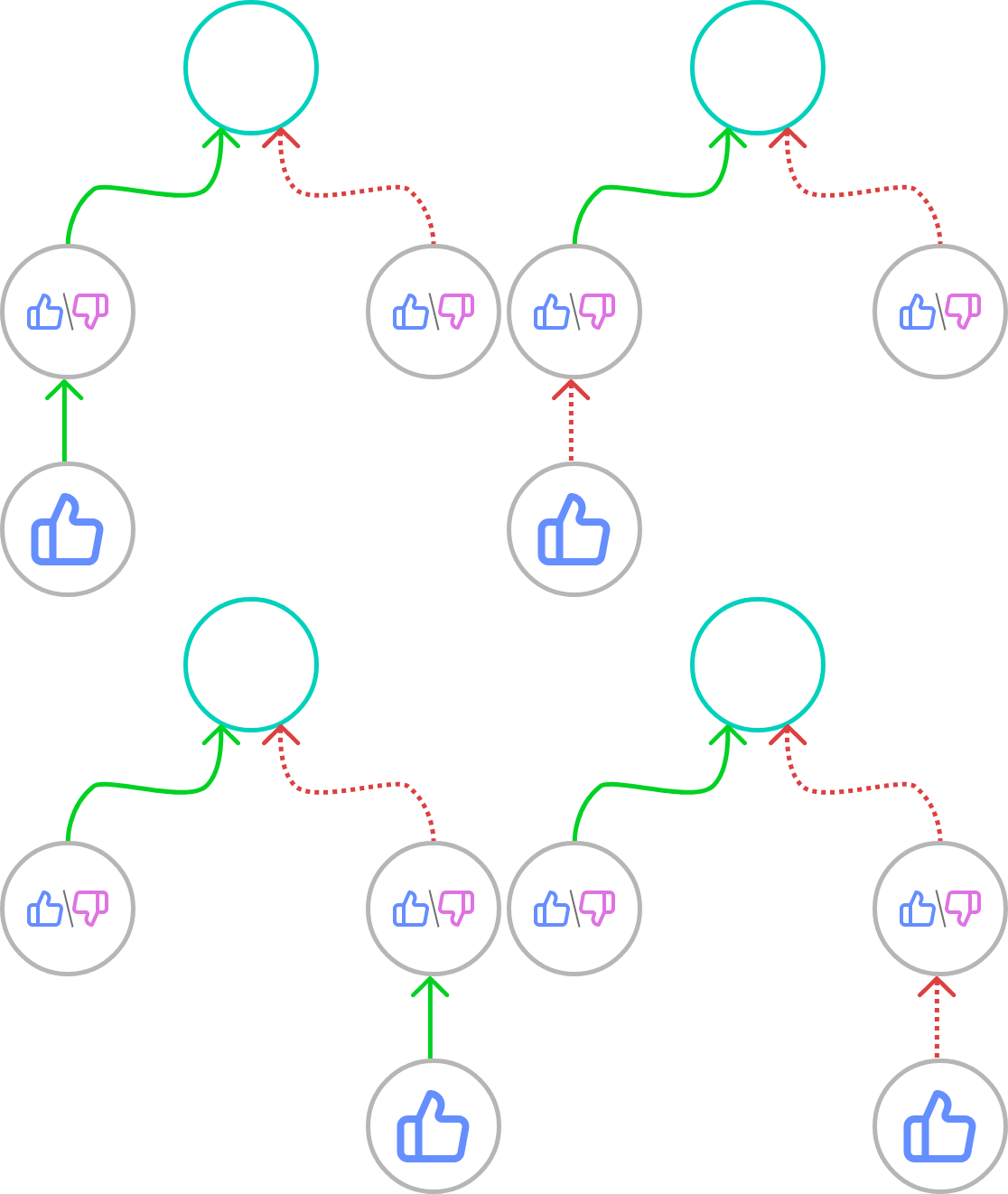}
    \caption{Abstract structure of the vote/depth complex debate variant.}
    \label{fig:vote-depth_complexity}
    
\end{figure}

\begin{figure}[htp!]
    \centering
    \includegraphics[width=0.5\linewidth]{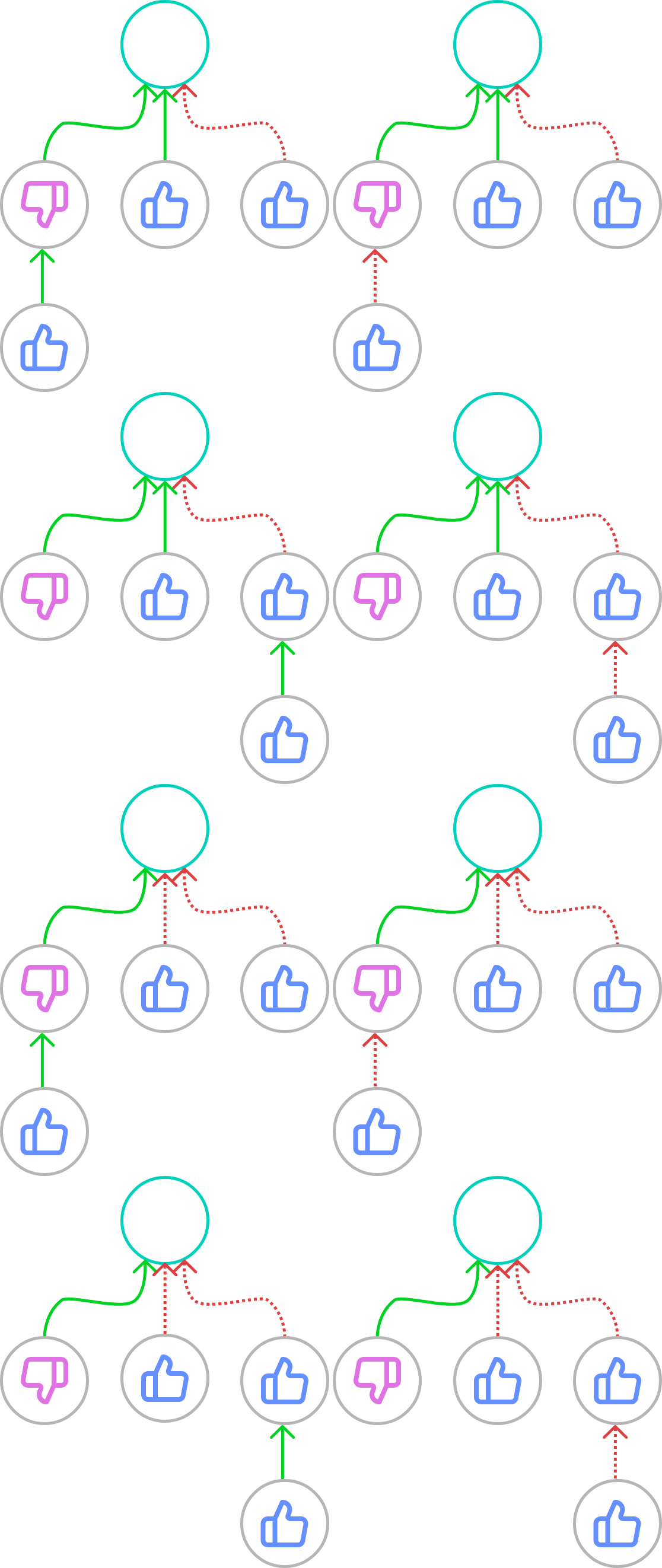}
    \caption{Abstract structure of the depth/breadth complex debate variant.}
    \label{fig:depth-breadth_complexity}
    
\end{figure}

\begin{figure}[htp!]
    \centering
    \includegraphics[width=0.5\linewidth]{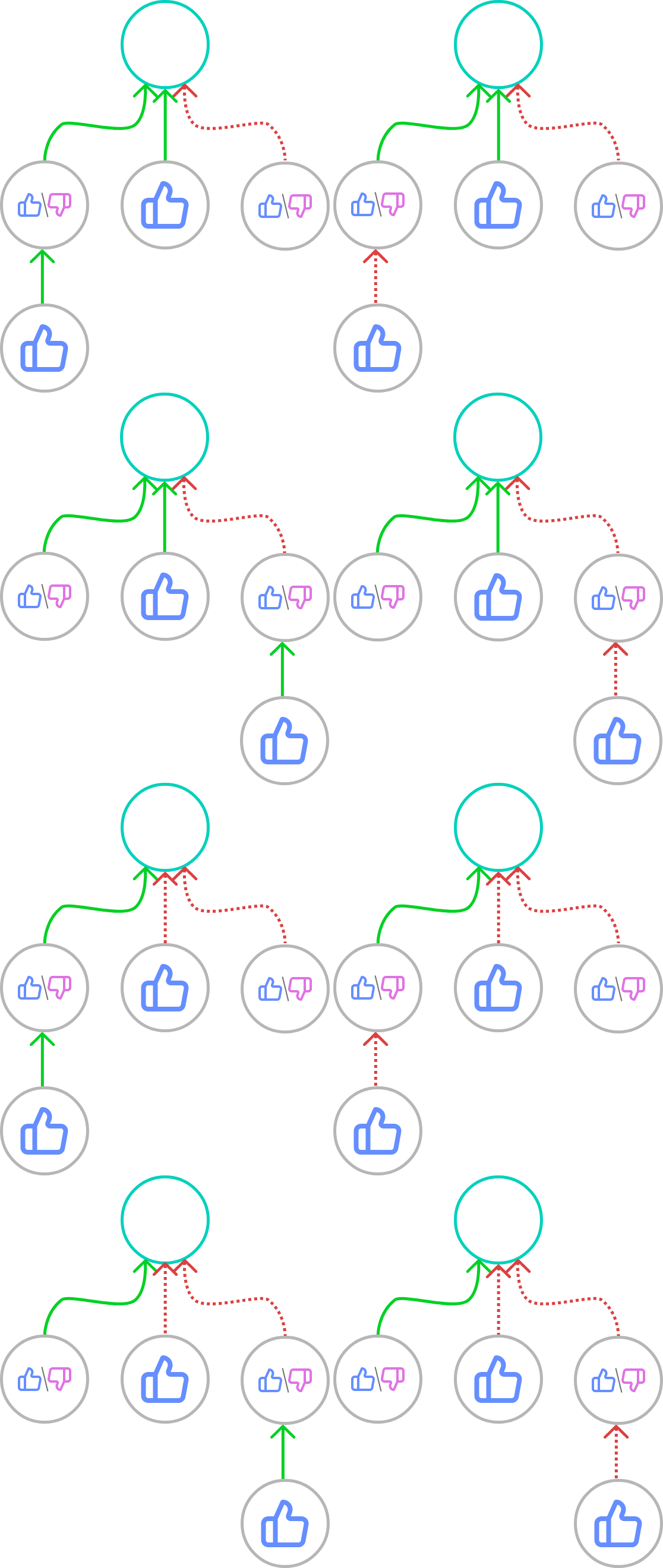}
    \caption{Abstract structure of the vote/depth/breadth complex debate variant.}
    \label{fig:vote-depth-breadth_complexity}
    
\end{figure}

In this section we represent the abstract structure of all debate variation complexities (see Figure~\ref{fig:simple}, \ref{fig:vote_complexity}, \ref{fig:breadth_complexity}, \ref{fig:depth_complexity}, \ref{fig:breadth_complexity}, \ref{fig:vote-breadth_complexity}, \ref{fig:vote-depth_complexity}, \ref{fig:depth-breadth_complexity}, and \ref{fig:vote-depth-breadth_complexity}).


\newpage
\section{Experimental Design}
\label{app:experimental-design}

The participants will be split such that 50\% of them assess simpler questions first and more complex questions after (Simple-Complex) and 50\% of them assess more complex questions first and simpler questions after (Complex-Simple). In both cases, users will be randomly allocated (50\%, 50\%)  which forecasting question they will assess (amongst Q1,Q2). For the 4 (variants of) debates the users will see, Alex’s forecast is also allocated randomly (1/3, 1/3, 1/3) amongst $<50\%,=50\%, >50\%$.

\begin{itemize}
    \item For the Simple-Complex case, they will be first shown the simple debate variant. We will then split the participants such that 1/3 assess vote complex debate variants, 1/3 assess breadth complex debate variants, and 1/3 assess depth complex debate variants.
    \begin{itemize}
        \item From vote complex debate variants participants will be split into 50\% vote/depth complex debate variants or 50\% vote/breadth complex debate variants.
        \item From breadth complex debate variants participants will be split 50\% vote/breadth complex debate variants or 50\%   depth/breadth complex debate variants.
        \item From depth complex debate variants participants will be split 50\% vote/depth complex debate variants or 50\%   depth/breadth complex debate variants.
        \begin{itemize}
            \item Then, regardless of what type of debate variants they get, they will be shown a vote/depth/breadth complex debate  variant.
        \end{itemize}
    \end{itemize}
    \item For the Complex-Simple case they will be shown the vote/depth/breadth complex debate variants. We will then split further the participants such that 1/3 assess vote/depth complex debate variants, 1/3 assess vote/breadth complex debate variants, and 1/3 assess depth/breadth complex debate variants.
    \begin{itemize}
        \item From vote/depth complex debate variants, participants will be split into 50\%  vote complex debate variants or 50\% depth complex debate variants.
        \item From vote/breadth complex debate variants, participants will be split 50\%  vote complex debate variants or 50\% breadth complex debate variants.
        \item From depth/breadth complex debate variants, participants will be split into 50\% depth complex debate variants or 50\% breadth complex debate variants.
        \begin{itemize}
            \item Then, regardless of what type of complex debate variant they get, they will be shown the simple variant.
        \end{itemize}
    \end{itemize}
\end{itemize}

\begin{figure*}[htp!]
    \centering
    \includegraphics[width=\linewidth]{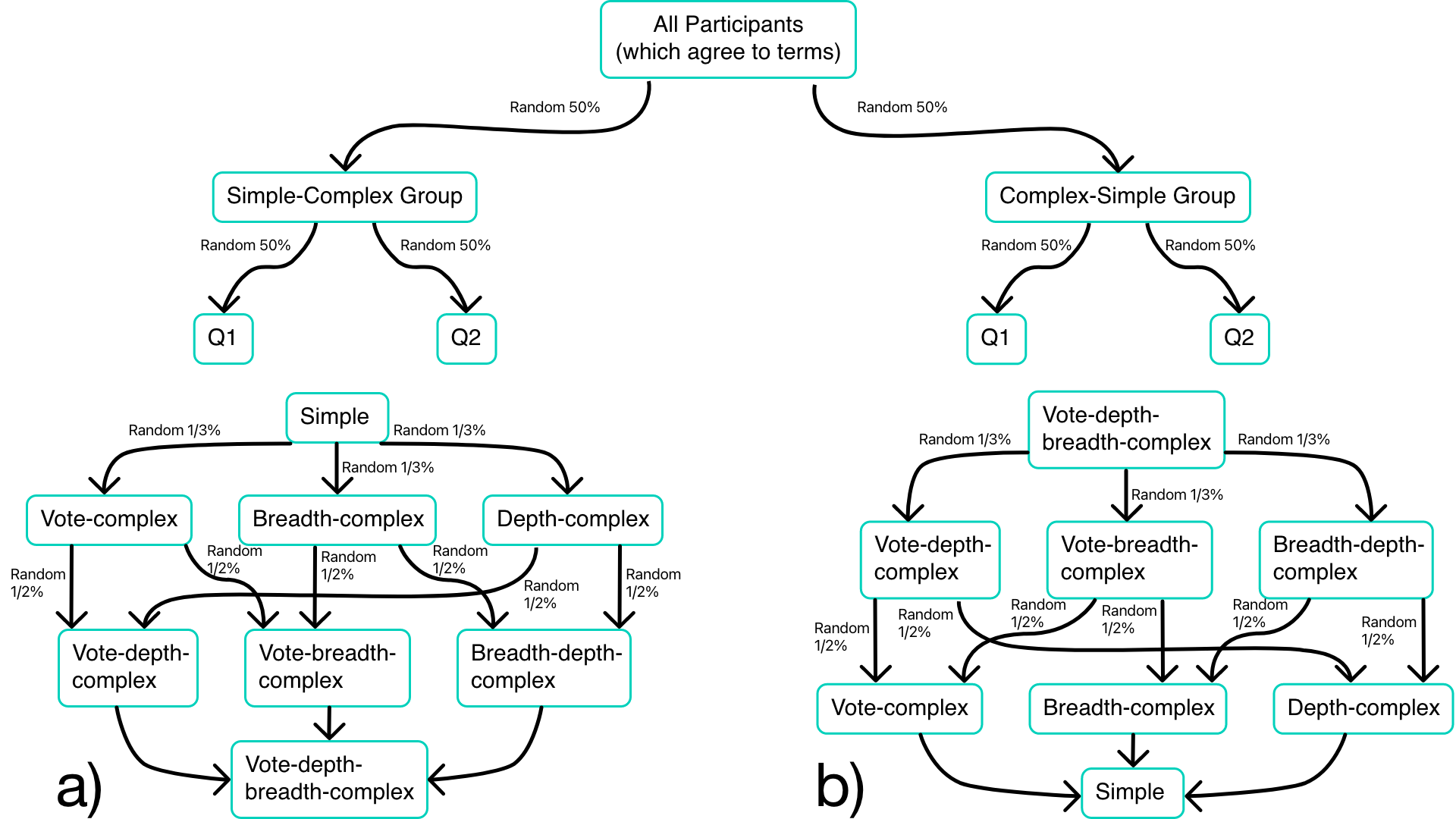}
    \caption{Experimental design for our user experiments where subfigure a) shows the experimental design for the Simple-Complex group and subfigure b) shows the experimental design for the Complex-Simple group.}
    \label{fig:experimental-design}
    
\end{figure*}

\end{document}